\documentclass[preprint,12pt,authoryear]{elsarticle}

\usepackage{amssymb}

\usepackage{multirow}

\usepackage{adjustbox}
\usepackage[table]{xcolor}
\usepackage{chngpage}
\usepackage{xcolor}
\usepackage{appendix}

\journal{Expert Systems with Applications}

\begin{document}

\begin{frontmatter}



\title{An Exploration of Features to Improve the Generalisability of Fake News Detection Models}


\author[inst1]{Mr. Nathaniel Hoy\corref{cor1}}
\ead{nathaniel.hoy2@brunel.ac.uk}

\author[inst1]{Dr. Theodora Koulouri}
\ead{theodora.koulouri@brunel.ac.uk}

\affiliation[inst1]{organization={Department Of Computer Science, Brunel University London},
            addressline={Kingston Lane}, 
            city={Uxbridge},
            postcode={UB83PH}, 
            state={Middlesex},
            country={United Kingdom}}

\cortext[cor1]{Corresponding author}

\begin{abstract}
Fake news poses significant global risks by influencing elections and spreading misinformation, making its detection a critical area of research. Existing approaches, primarily using Natural Language Processing (NLP) and supervised Machine Learning, achieve strong results under cross-validation and hold-out testing but struggle to generalise to other datasets, even those within the same domain. This limitation stems from reliance on coarsely labelled training datasets, where articles are often labelled based on their publisher, introducing biases that token-based representations such as TF-IDF and BERT are sensitive to. While Large Language Models (LLMs) represent a promising development in NLP, their application to fake news detection remains limited. This study demonstrates that meaningful features can still be extracted from coarsely labelled datasets to improve model robustness for real-world scenarios. Stylistic features, including lexical, syntactic, and semantic attributes, are explored as an alternative due to their reduced sensitivity to dataset biases. In addition, novel `social-monetisation' features are introduced, capturing economic incentives behind fake news, such as the presence of advertisements, external links, and social media sharing elements. The study employs the coarsely labelled NELA 2020-21 dataset for training and the manually labelled Facebook URLs dataset for external validation, representing a gold standard for evaluating model generalisability. The results highlight the limitations of token-based models, when trained on coarsely labelled data. Additionally, this study contributes to the limited evidence on the performance of LLMs such as LLaMa in this domain. The findings indicate that stylistic features, complemented by social-monetisation attributes, provide more generalisable predictions for real-world scenarios in comparison to token-based methods and LLMs. Statistical and permutation feature importance analyses further reveal the potential of these features to enhance performance and address dataset biases, offering a path forward for improving fake news detection models.
\end{abstract}

\begin{highlights}

\item Demonstrated poor generalisability of token-representations/LLMs on real-world data

\item Established stylistic features lead to more generalisable and balanced models

\item Proposed social-monetisation features increased accuracy across datasets

\item Established a simplified feature-set matching comprehensive set's performance

\end{highlights}

\begin{keyword}
fake news \sep misinformation \sep machine learning \sep natural language processing \sep feature engineering \sep generalizability
\end{keyword}

\end{frontmatter}



\section{Introduction}
Fake news has been a topic of interest since the term was popularised at the time of Trump’s 2016 Presidential Election bid. It is typically characterised as content that appears to be news but is intentionally misleading for the purposes of generating profit through advertising or exerting political influence \citep{Allcott2017}. In recent times it has maintained its relevance in public discourse, particularly with the rise of generative text models that are capable of generating large amounts of misinformation quickly and easily \citep{Xu2023}. Detecting fake news before dissemination is crucial to uphold information integrity and maintaining public trust in media and institutions. In democratic societies, fake news can manipulate public opinion, sway elections, and undermine governance \citep{Morgan2018}. Moreover, false information about health crises, emergencies, or scientific discoveries can endanger public safety \citep{Nelson2020}. Socially, it deepens divisions, fuels polarisation, and exacerbates societal tensions \citep{Olan2024}. Therefore, effective detection and mitigation of fake news not only protects individuals from harm but also upholds the essential principles of truth, transparency, and responsible communication necessary for a well-functioning society.

Researchers in the field of Computer Science have aimed to address this problem, often  using Natural Language Processing (NLP) and Machine Learning (ML) classification techniques. While such approaches report good results under cross-validation and holdout test conditions, evidence suggests that current approaches struggle to generalise, particularly when relying on token-representation methods such as Bag-of-Words (BoW), Term Frequency - Inverse Document Frequency (TF-IDF) and Bidirectional Encoder Representations from Transformers (BERT). This issue is further exacerbated by the use of coarsely labelled datasets, commonly used in the literature, that label articles based on their publisher rather than individually through manual fact-checking. While Large Language Models (LLMs) have emerged as transformative tools in NLP, enabling sophisticated and context-aware text analysis, their application to fake news detection remains limited. Given these issues, investigating the issue of generalisability in both token-based and LLM-based fake news detection models is the primary focus of this study.

Evidence of the poor generalisability of fake news detection models using token-representations can be found in \cite{Alnabhan2024}, which explored the use of deep learning models and embeddings such as Glove and BERT. Results from this study demonstrate that models achieve high accuracies under holdout test conditions (around 99\%) but suffer a drop in accuracy of around 30\% when evaluated on several different datasets. Similar results are also observed in \cite{Hoy2022}, where models trained on one political news dataset, struggle to generalise to other political news datasets (also observing a ~30\% drop in accuracy). \cite{Hoy2022} additionally demonstrate, through the use of the Local Interpretable Model-agnostic Explanations (LIME) package in Python, that token-representations were more sensitive to topical biases within the datasets on which the models were trained, while stylistic features, though still performing poorly in terms of generalisability, demonstrated less sensitivity to such topical biases. These findings also indicate the need for more robust testing methodologies and datasets to ensure that models perform as expected in the real-world. 

Motivated by these findings, this study seeks to enhance the generalisability of fake news detection models by identifying a set of features that exhibit reduced sensitivity to biases present within datasets. Building on preliminary findings hinting at the lower sensitivity of stylistic features to these biases, these features will serve as the basis for further exploration. Additionally, aligning with approaches observed in existing literature, the study will propose additional features independent to the article text. This exploration aims to determine if these features contribute to the development of more generalisable fake news detection models—that is, whether they perform well when tested on different datasets— compared to the current approaches that largely rely on token-representations. Owing to the limited evidence of the efficacy of LLMs in this domain, the also study includes an evaluation of LLaMa to provide a basis for comparison against other approaches in addressing these challenges. Notably, this study employs the Facebook URLs dataset as the external validation set to evaluate generalisability; this dataset is not publicly available and is unique in that it has been manually labelled by an external fact-checking organisation, making it more representative benchmark than other datasets used in the literature. 

To meet the aims outlined above, this paper will commence with an exploration of the relevant literature; in particular, Section 2 will discuss studies that have focused on the use of stylistic features and their efficacy, arguing that the use of stylistic features may be more effective in producing more generalisable models that are less sensitive to topical biases. Building on this exploration, three research questions pertaining to this study will be formulated. Section 4 will discuss the methodology of the experimental work designed to address the research questions. It will commence with the methods underpinning the experiments, including the data collection process, feature-sets, training and testing methodology, machine learning algorithms and evaluation metrics chosen for the experiments. Section 5 will detail the results of the experiments. Section 6 will discuss these results in relation to existing research, motivating the paper’s conclusions and proposed directions of future work, which are outlined in Section 7.

\section{Related Work}
The challenge of detecting fake news has prompted extensive research across various domains, leading to the development of numerous models and approaches. This section reviews key studies that have explored token-level representations, stylistic features, and multimodal approaches to enhance the accuracy and generalisability of fake news detection models

\subsection{Overview}

Current approaches to fake news detection largely focus on using a variety of supervised ML algorithms and features. Typically, such approaches are tested using either holdout testing or cross-validation, using an unseen portion of the datasets on which they were trained. Overall, current approaches achieve accuracies of ~80\% on average, with many achieving significantly higher results \citep{Hoy2021}. The features that are used in such approaches can broadly be categorised as either: (i) content-based, which are features derived from things such as article text, title, publisher and images; (ii) socially-based, which are features derived from social networks data (typically from X/Twitter), such as the relationships between users who share fake news and their profiles; and (iii) feature fusion, which includes a combination of these features \citep{Xie2020}. Of these groups, content-based features are overwhelmingly the most popular, with many methods having a particular focus on features derived from the article text. As social media companies such as X/Twitter become more restrictive in the access provided to their social networks through their APIs \citep{Blakey2024}, it is likely that content-based features will remain the most dominant approach in the field of fake news detection, making this category of approach the most fruitful to study. 

\subsection{Content-Based Textual Features}

Of the sub-categories of content-based features, textual features are the most prominently used. These can be broadly divided into the following sub-categories:

\begin{itemize}
    \item \textbf{Token-Level Representations:} These convert words into numerical vectors using methods like Bag of Words (BoW), TF-IDF, and word embeddings such as Word2Vec and BERT \citep{Thota2018}. BoW and TF-IDF represent text data as vectors based on word frequency and importance, respectively, while word embeddings like Word2Vec and BERT capture semantic relationships and contextual information between words. Studies such as \cite{Kaur2020} and \citep{Poddar2019} have shown high accuracies using these methods with various machine learning models, demonstrating their effectiveness in the field of fake news detection under hold-out testing conditions

    \item \textbf{Stylistic Features:} These include statistical features (e.g., average word length, sentence complexity, part-of-speech tags) and psycholinguistic features (e.g., sentiment analysis scores) generated by analysing the text corpus. Tools like Linguistic Inquiry Word Count (LIWC) produce such features, encompassing linguistic aspects and psycholinguistic processes. These features provide insights into the author's writing style and emotional tone, enhancing the analysis of text data \citep{Gravanis2019,Spezzano2021}. \cite{Fernandez2019} demonstrated strong performance using stylistic features, achieving up to 94.2\% accuracy in their experiments, highlighting the importance of stylistic analysis in improving model performance
\end{itemize}

Studies may also explore the combination of token-level representations and stylistic features to leverage the strengths of both approaches. Research combining these features have demonstrated strong performance in fake news detection. For instance, \cite{ngada2020fake} demonstrated over 95\% accuracy on the Kaggle `fake and real news' dataset across six different classification algorithms by integrating token-level and stylistic features. Similarly, Verma et al. (2021) showed that such combined features can generalise well across four datasets. However, these generalisation results stand in significant contrast to the wider literature \citep{Hoy2022,Gautam2020,Blackledge2021,Janicka20191089,Alnabhan2024,Liu2024}, suggesting a likely overlap between the datasets used, particularly in experiments leveraging a Kaggle dataset, where data collection methodologies are not often described and combining datasets for training and testing is a common practice. Such overlap may artificially inflate performance metrics and does not reflect the challenges of generalising to entirely unseen data. These findings underscore the need for careful dataset selection and validation to ensure meaningful evaluations of generalisability.

Beyond traditional token-based and stylistic methods, Large Language Models (LLMs) such as GPT and LLaMA represent a distinct advancement in natural language processing. Unlike embedding-based approaches like BERT, which focus on generating contextual embeddings for token-level representations, LLMs leverage extensive pretraining on vast datasets to perform text classification tasks holistically. These models excel in zero-shot and few-shot learning scenarios, demonstrating strong performance across tasks such as sentiment analysis and summarisation without requiring extensive fine-tuning \citep{kojima2022large}.

Preliminary evidence suggests, however, that LLMs struggle to achieve similar success in the fake news detection domain. Unlike their strong performance on tasks such as summarisation or sentiment analysis, their results in fake news detection have been consistently poor. Recent studies, including \cite{pavlyshenko2023analysis}, highlight significant differences in performance between LLMs and specialised NLP classifiers that leverage token-based features, with the latter consistently outperforming LLMs in this domain.

\subsection{Multimodal Appraoaches using Textual Features}

Multimodal approaches leverage various types of data, including textual features, to enhance the performance of fake news detection models. By combining different data sources, these approaches aim to create more comprehensive and accurate models

One strategy involves combining LIWC with other features. Tools like LIWC are often used in conjunction with additional data to improve model accuracy. \cite{Ahmad2020} and \cite{Shu2019} demonstrated that integrating LIWC features with user profile data results in high accuracies across various datasets. This combination allows models to capture not only the linguistic and psycholinguistic aspects of the text but also contextual information from user profiles

Another strategy is the integration of text and image data. \cite{Spezzano2021} highlighted the benefits of combining textual features with image data. By integrating LIWC features with visual information, their study showed significant improvements in model performance. This approach leverages the strengths of both textual and visual analysis, providing a more holistic view of the content

\subsection{Current Issues in Fake News Detection}

Despite positive outcomes in many studies, issues remain with datasets used for training fake news detection models, particularly those relying on token representations. Dataset size is a critical issue, as collecting and accurately labeling a large number of news articles is challenging \citep{DUlizia2021}. To manage this, articles are often labeled based on their publisher as a proxy for accuracy, which can introduce topical biases \citep{Torabi2019}. This can lead to models that perform well in hold-out test conditions but struggle to generalise outside of the training dataset \citep{Suprem2022}

Limited studies have explored the generalisability of these models. \cite{Gautam2020} observed a 39\% accuracy drop when models trained on political news were tested on celebrity news. Similarly, \cite{Castelo2019} found comparable accuracy drops across different news domains using small datasets of fewer than 500 articles. Multimodal approaches, like those examined by \cite{Liu2024}, combining text and image features, showed some cross-domain generalisation but significant accuracy drops to around 55\% under cross-dataset conditions. Notably, embeddings from more recent large language models such as GPT-4 also suffer from an inability to generalise when trained on these coarsely labelled datasets, as demonstrated by \cite{Alnabhan2024}

\cite{Hoy2022} focused on four political news datasets to assess generalisability within a single domain. They found significant accuracy drops and sensitivity to biases in token-representation models. Interestingly, while stylistic features also struggled with generalisability, they performed more consistently across datasets, suggesting lower sensitivity to dataset biases

This motivates further exploration of stylistic features, forming the foundation of the research presented in this paper. Studies by \cite{Shu2019BeyondCont} and \cite{Spezzano2021} support incorporating supplementary features beyond the text, arguing that combining different feature categories outperforms single-category approaches. Similarly, \cite{Liu2024} highlights the need to explore additional features to create more generalisable models, especially in cross-dataset conditions

\subsection{Proposed Features}
Given this evidence, this study seeks to propose a set of four novel features with the goal of producing more generalisable fake news detection models. These features are outlined as follows:

\begin{itemize}
    \item \textbf{Frequency of Ads:} One of the primary motivations behind the creation and dissemination of fake news is financial gain through advertising. According to \cite{Allcott2017}, fake news websites often rely on sensationalist and misleading content to attract high volumes of traffic, which in turn increases their advertising revenue. These sites typically feature a large number of advertisements, as their business model is heavily reliant on generating ad impressions and clicks. Therefore, the number of adverts associated with a given article could be a significant indicator of fake news. Articles that contain an unusually high number of ads may be designed to maximise revenue rather than to provide factual information, making this a critical feature to include in fake news detection models

    \item \textbf{External Links:} Similar to advertising, the prevalence of external links in an article can also be an indicator of fake news, especially when these links are intended for affiliate marketing purposes. Fake news articles often include numerous external links that direct readers to other sites, which can generate affiliate income for the publisher each time a link is clicked. This tactic is particularly common in misinformation related to healthcare and other high-interest topics, as noted by \cite{Rehman2022}

    \item \textbf{Social Media Share Links:} The role of social media in the spread of fake news is well-established, with platforms like Facebook and X/Twitter being primary channels for misinformation dissemination. One of the mechanisms that facilitate this spread is the use of visual cues, such as share buttons, which prompt habitual behaviour in social media users \citep{Ceylan2023}. When users encounter these visual cues, they are more likely to share the content without critically evaluating its veracity. Including 'call to action' links that lead to social media platforms in the analysis is essential, as these links can significantly amplify the reach of fake news articles. By encouraging readers to share content on social media, these articles can quickly go viral, spreading misinformation at an unprecedented rate. Therefore, factoring in Facebook and X/Twitter links is expected to be important in identifying articles that are designed to exploit social media behaviour for rapid dissemination. It is important to note these social media features are distinct from others seen in the literature, which typically focus on user profiles and relationships between tweets and users
\end{itemize}

This novel group of features shall be characterised as ‘social-monetisation’ features

\section{Research Questions}
The analysis in Section 2 provides the motivation for a focused investigation into the use of generalisable stylistic features as well as the novel social-monetisation features that may lend themselves to improved generalisability (the ability of a model to perform well when tested on a different dataset than the one on which it was trained) of fake news detection models. As such, the objectives of this study can be formalised into the following three research questions:

\begin{itemize}
    \item \textbf{RQ1.} How well do fake news detection methods using token-\\representations/LLMs generalise? 
    \item \textbf{RQ2.} Do fake news detection methods using stylistic features generalise better than fake news detection models using token-representations/LLMs?
    \item \textbf{RQ3.} Do fake news detection methods using stylistic features and the proposed social-monetisation features generalise better than models using stylistic features only?  
\end{itemize}

\section{Methodology}
\begin{figure}[h]
\centering
\includegraphics[width=\linewidth]{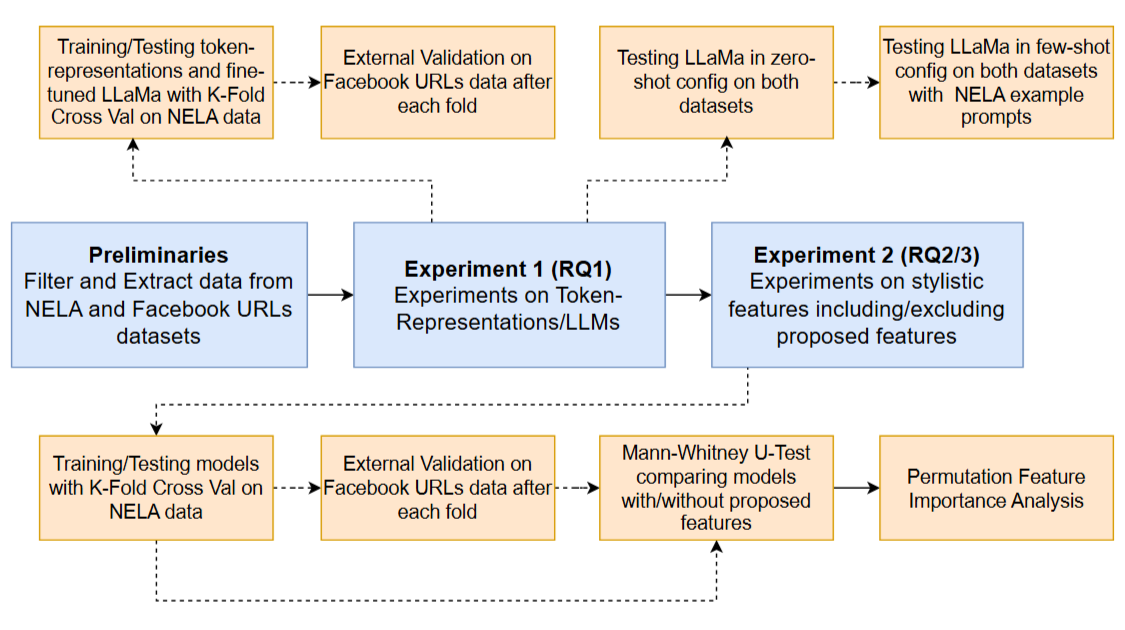}
\caption{Study Overview}
\label{fig:overview}
\end{figure}

This section details the methodology used to address the research questions outlined in Section 3. The study conducted two experiments (summarised in Figure 1) to achieve this. 

The first experiment aimed to evaluate the generalisability of commonly used token-representations, including Bag of Words (BoW), TF-IDF, Word2Vec and BERT, as well as the Large Language Model `LLaMA', addressing RQ1. Models using token-representations were trained on the NELA 2020-21 dataset and their performance evaluated using K-fold cross-validation. For LLaMA, the zero-shot configuration relied on its pre-trained knowledge and structured prompts, while the few-shot configuration incorporated labelled examples from the NELA dataset to guide classification. The fine-tuned LLaMA model followed a methodology similar to the token-based approaches, enabling a direct comparison of generalisability across techniques. Models trained/fine-tuned on the NELA dataset were then evaluated using external validation on the Facebook URLs dataset. By evaluating token-representations and LLMs in this manner, the study established a baseline for comparison with other feature-sets used in the second experiment

The second experiment focused on assessing the generalisability of five groups of stylistic features and evaluating the impact of newly proposed social-monetisation features. This experiment addressed two research questions: the effectiveness of stylistic features (RQ2) and the improvement in model performance through the inclusion of social-monetisation features (RQ3). Similar to the first experiment, models were trained using the NELA 2020-21 dataset, initially evaluated using K-fold cross-validation, before being externally evaluated on the Facebook URLs dataset to assess generalisability. These tests were conducted, both with the inclusion and exclusion of the proposed social-monetisation features. This was done to determine if the proposed features resulted in a statistically significant improvement in generalisability performance using the Mann-Whitney U-Test. 

The following subsections elaborate on the methodology. Section 4.1 outlines the data collection process and resulting datasets that were used in the experiments. Section 4.2 to 4.3 describe the features that were extracted from these datasets in relation to Experiment 1 and Experiment 2 respectively. Section 4.4 outlines the machine learning algorithms that were used in these two experiments and how they were trained and tested. 

\subsection{Data Collection \& Processing}

This section outlines the datasets and data extraction methods used. Owing to the nature of the proposed social-monetisation features, the dataset required the source URL of the articles to facilitate the extraction of these features. \cite{Capuano2023}'s systematic review lists several datasets used in content-based fake news detection. However, out of the 19 datasets reviewed, only three — FakeNewsNet, Buzzfeed, and Celebrity fake news—include the article's source URL. These datasets are relatively small, which limits the likelihood of producing a generalisable model. To develop a more comprehensive and reliable model, a larger dataset is necessary. Therefore, the NELA series of datasets was chosen for its large size and inclusion of article URLs, providing a more extensive and diverse data source for training. Using a dataset of this size also ensures that a significant number of articles can be extracted to compensate for pages that are no longer available. While not as frequently used in the literature, a number of studies make use of this dataset including \cite{Horne2020}; \cite{Raj2023} and \cite{Raza2022}

The latest iterations of this dataset released in March 2023, NELA 2020 and 2021, were chosen for this study. Each dataset contains over a million articles from various sources and are coarsely labelled, with each article's legitimacy derived from its source's aggregated label from seven assessment sites: Media Bias Fact Check, Pew Research Center, Wikipedia, OpenSources, AllSides, Buzzfeed News, and Politifact. The labels are categorised as unreliable, mixed, and reliable. For this study, only `unreliable' and `reliable' labels were used, excluding the `mixed' label to align with the binary labels in the external validation dataset

\begin{figure}[h!]
\centering
\includegraphics[scale=0.7]{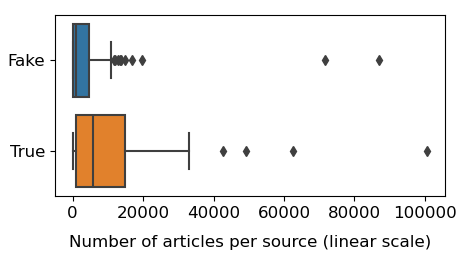}
\caption{Articles per Source Prior to Extraction}
\label{fig:art-per-src-pre}
\end{figure}

\begin{figure}[h!]
\centering
\includegraphics[scale=0.7]{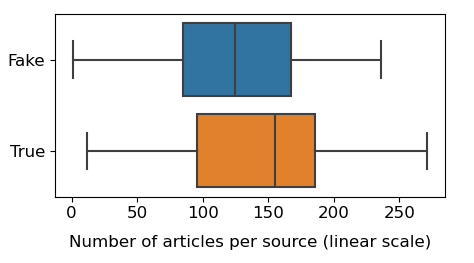}
\caption{Articles per Source Post Extraction}
\label{fig:art-per-src-post}
\end{figure}

The combined NELA 2020-21 dataset includes 3,635,636 records from 525 unique sources. After joining the labels file and excluding the `mixed' category, the dataset consists of 1,013,808 `true' and 551,051 `fake' articles from 224 sources. To prevent any single source from dominating the training set \ref{fig:art-per-src-pre}, the number of URLs extracted from each source was reduced using the 1st quartile as a threshold (285 articles per source), resulting in a final set of 22,230 `true' and 25,650 `fake' articles from 168 sources (Figure \ref{fig:art-per-src-post}).

The Facebook URLs Dataset was chosen as an external validation dataset owing to its unique position as a dataset collected in a ‘real-world’ context and granular labelling by a third-party fact-checking organisation. Its individual article labels provide a robust standard for assessing model accuracy and practical applicability in fake news detection. This stands in contrast to commonly used datasets in the field, which often employ coarse labels based on article publishers, potentially misrepresenting the true nature of fake news. By using a coarsely labelled dataset for training and a manually labelled dataset for testing, the aim is to demonstrate that despite the limitations of coarsely labelled datasets, meaningful features can still be extracted to develop robust models applicable in real-world scenarios

The Facebook URLs dataset contains over 38 million URLs shared on Facebook since January 1, 2017, with 35,924 records identified as fake news. The dataset is protected with differential privacy, ensuring no information can be gathered regarding individuals \citep{Messing2020}. Given its restricted accessibility and limited usage in prior studies, this research represents one of the few to utilise the Facebook URLs Dataset for fake news classification, following a study by \cite{Barnabo2022}. The dataset initially comprised 28,271 fake and 7,653 true records, with non-English articles filtered out based on `US' and `UK' values in the `Public Shares Top Country' field, resulting in 14,354 fake and 1,468 true records. To enhance dataset quality, URLs referring to Tweets and videos were excluded. Class balancing was implemented during experimentation. Due to its size, the Facebook URLs Dataset served as a test set for cross-dataset testing, complementing the larger training datasets to bolster the model's generalisability and validate its performance in diverse real-world scenarios

In order to extract the raw textual data from the URLs in these datasets, the BeautifulSoup library was used. As many webpages in these datasets may no longer be available, particularly in relation to ‘fake’ news pages, initial extraction was attempted through the use of the Wayback Machine API (Internet Archive). This was done to increase the likelihood of extracting a webpage with a complete article and not a splash page indicating the article had since been deleted. In instances where webpages were not available in this archive, a final extraction attempt was made directly from the webpage using the URL provided in the dataset to account for cases where webpages may not yet have been added to the Internet Archive. If through these methods a complete article was not extracted, the URL would be excluded from the resulting dataset. 

In cases where full articles were available, rather than attempt to accurately extract only the text pertaining to the news articles from these URLs, all textual elements are extracted from the body of the webpage. While this may introduce additional noise to the feature-sets, it was a deliberate choice. Websites have different layouts, styles and coding structures, making it challenging to consistently and accurately extract only the article text. It is argued that models that extract all textual elements from the webpage body are more adaptable to the varying structures and formats of webpages and, as such, have the potential to be more robust and scalable across a wider range of online content. Following this data extraction phase, pages returning \textless3KB of data were excluded, as it was observed that pages with less than this amount of data had typically had their articles removed. The resulting datasets are summarised in Table \ref{tab:Dataset Statistics}:

\begin{table}[h]
\centering
\caption{Dataset Statistics}
\label{tab:Dataset Statistics}
\vspace{5pt}
\begin{tabular}{|l|c|c|}
\cline{2-3}
\multicolumn{1}{c|}{} & \textbf{NELA 2020-21} & \textbf{Facebook URLs Dataset} \\
\multicolumn{1}{c|}{} & \textbf{(Training Dataset)} & \textbf{(External Validation)} \\
\hline
\textbf{Fake} & 10,529 & 5,355 \\
\textbf{True} & 10,487 & 798 \\
\hline
\end{tabular}
\end{table}

\subsection{Experiment 1 Features: Token-Representations \& LLMs}

\begin{figure}[ht]
\centering
\includegraphics[scale=0.37]{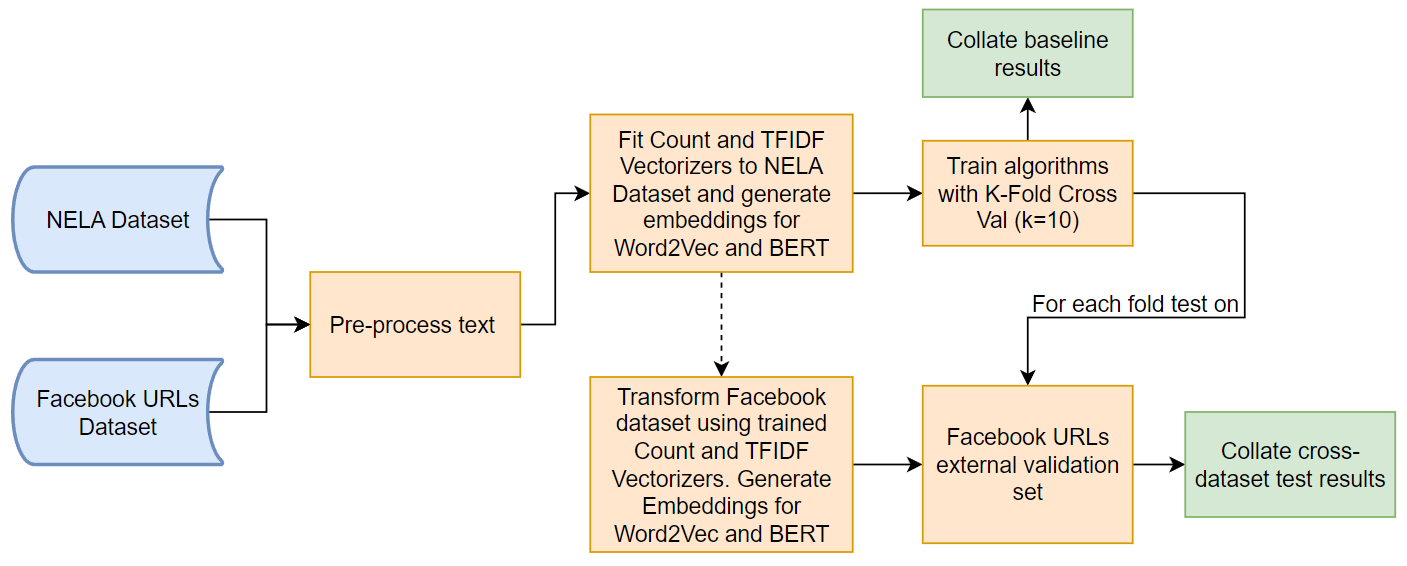}
\caption{Experiment 1 - Token-Representations Overview}
\label{fig:exp1}
\end{figure}

This section outlines the features to be used in the first experiment, addressing RQ1. This experiment aims to address RQ1, by exploring how well models using token-representations generalise between two different datasets of the same topic, using the NELA and Facebook datasets. An overview of the procedure followed in this experiment is provided in Figure \ref{fig:exp1}. In this section, each of the token-representations used in this experiment and the libraries used in extracting these features from the datasets are outlined. The results of this experiment are presented in Section 5.1. 

\subsubsection{Token-Representations}

The token-representations chosen for this study are motivated by the systematic review by \cite{Capuano2023}, which identified the following token-representations among the most popular in the literature for content-based fake news detection:

\begin{itemize}
    \item \textbf{Bag-of-Words (BoW)} converts text into fixed-length vectors based on word frequencies. It does not take into account word-order. In this study, BoW was implemented with SKLearn's CountVectorizer with a max\_features parameter of 10,000 words. 

    \item \textbf{TF-IDF} enhances BoW by considering word frequency relative to the dataset, capturing token significance. This technique, therefore, better captures the significance of tokens compared to BoW. However, like BoW of words it fails to account for word-order. TF-IDF is among the most popular feature extraction approaches, as identified by \cite{Capuano2023}, owing to its simplicity and focus on informative tokens. In this study, TF-IDF was implemented with SKLearn's TFIDFVectorizer and max\_features set to 10,000 words

    \item \textbf{Word2Vec} generates word embeddings by training a neural network to predict target words from their context, capturing semantic meaning but ignoring word order (similar to the previous two techniques). It is one of the most popular feature extraction methods, identified in approximately 25\% of studies in an ongoing systematic review by the authors. For this study, a pre-trained Word2Vec model trained on Google News data was implemented using the Zeugma library

    \item \textbf{BERT}, similar to Word2Vec, also generates word-embeddings but with the distinct advantage of being context-dependent thus allowing for unique representations of words morphologically similar words. BERT achieves this through a novel approach of training at the sub-word level, encoding word-positions and training on tasks such as Masked-Language-Modelling (MLM) and Next-Sentence Prediction (NSP). Although other, more advanced,  transformer-based models like GPT-4 exist, they are still much less well-established in the literature for this specific task. Therefore, BERT has been chosen for this study In this study, BERT was implemented using the SentenceTransformers library using the pre-trained ‘bert-base-uncased’ model. Additionally, a fine-tuned version of this model trained on the NELA dataset was employed to further optimise performance for the fake news detection task.
\end{itemize}

In the case of the BoW and TF-IDF approaches, the following steps were taken to remove any unwanted noise from the text: (i) converting the text to lowercase to ensure all words were treated uniformly; (ii) lemmatising the text; and (iii) removing punctuation, URLs, Twitter handles, extra-whitespace and stop words. Word2Vec and BERT did not undergo the above three steps  because these techniques require contextual information in order to generate their embeddings. 

\subsubsection{LLMs}

Building on the foundation of traditional token-based features, this study also incorporates the large language model \textbf{LLaMa 3.2-1B}, a transformer-based model with 1 billion parameters. Chosen for its robust natural language understanding and computational efficiency, the model provides a strong baseline for evaluating advanced detection techniques.

LLaMa was utilised in three configurations: zero-shot, few-shot, and fine-tuning. In the zero-shot configuration, the model leveraged its pre-trained knowledge without task-specific training, providing an initial benchmark for its capabilities. Few-shot learning introduced a small number of labeled examples to guide predictions, while fine-tuning adapted the model comprehensively by training it on labeled datasets.

To operationalise LLaMa in zero-shot and few-shot learning, structured prompts were designed to align with the task requirements. The system prompt defined the classification task and response format, while user prompts provided the input text for classification. For example:

\begin{verbatim} 
{"role": "system", "content": 
"Classify the text as `fake' news or `real' news. 
Respond only with `fake' or `real'."} 

{"role": "user", "content": 
"Text: {INSERT ARTICLE TEXT} 
\n Is this `fake' or real' news?"} 
\end{verbatim}

In the few-shot configuration, a series of four labelled examples (two from each class: fake and real) were provided as prompts to demonstrate the classification task to the model. The subsequent prompts iteratively processed the remaining dataset, excluding the initial examples, to classify each instance while leveraging in-context learning for improved performance. This approach was adopted to utilise the model’s capability to learn task-specific patterns from minimal examples while preserving token space for processing longer texts effectively.

For fine-tuning, \textit{Low-Rank Adaptation (LoRA)} and \textit{Parameter-Efficient Fine-Tuning (PEFT)} were employed \citep{pavlyshenko2023analysis}. LoRA targeted the transformer’s query (\texttt{q\_proj}) and value (\texttt{v\_proj}) projection layers with a rank of 8, alpha scaling of 32, and dropout of 0.1, enabling efficient learning of task-specific patterns. PEFT further optimized this process, allowing the model to retain its pre-trained knowledge while adapting to the binary classification task.

The fine-tuning process included tokenizing input texts to a maximum length of 512 tokens, configuring training with a learning rate of $2 \times 10^{-5}$, weight decay of 0.01, and three epochs, and managing the process using the Hugging Face \texttt{Trainer}. This configuration complemented the zero-shot and few-shot setups by embedding task-specific knowledge directly into the model, enabling scalable and efficient fake news detection.

\subsection{Experiment 2 Features: Stylistic \& Proposed Social-Monetisation \\Features}

\begin{figure}[ht]
\centering
\includegraphics[scale=0.37]{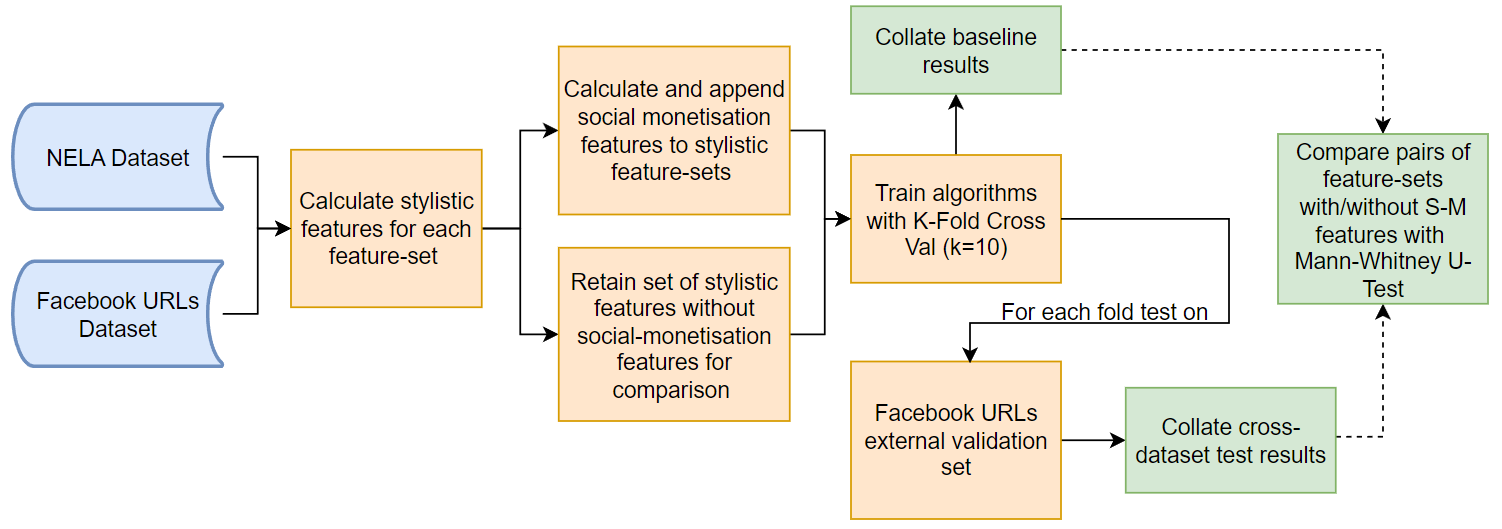}
\caption{Experiment 2 - Stylistic Features Overview}
\label{fig:exp2}
\end{figure}

The following section outlines the stylistic features that were used in addressing RQ2 Experiment 2 follows a similar structure to Experiment 1 and evaluates the generalisability of five groups of stylistic features proposed by previous research and compares it against the results of Experiment 1, which explored the generalisability of token-representations. Then, it addresses RQ3 by exploring whether the four social-monetisation features proposed by this study improve generalisability. An overview of the procedure followed in Experiment 2 is shown in Figure \ref{fig:exp2}. The stylistic features and social monetisation features used in the experiment are presented in Sections 4.3.1 and 4.3.2, respectively. The results of this experiment are described in Section 5.2.

\subsubsection{Selection of Stylistic Features}

Five groups of stylistic features, proposed in the literature, are evaluated in this study. Each feature group varies in complexity, with the first group focusing solely on linguistics, while subsequent feature-sets progressively incorporate additional groups such as psycholinguistics and document complexity. Due to the diverse scales of many features within these groups, we applied StandardScaler to ensure uniform treatment by machine learning algorithms. The selection of these five groups was motivated by their inclusion in the limited number of studies exploring generalisability of fake news detection models (with the exception of the NELA feature-set which was chosen owing to the use of the NELA dataset in this study). However, these studies only used coarsely labelled datasets for external validation. As such, we aimed to observe their performance using real-world data from the Facebook URLs dataset, underscoring the relevance of these features in practical applications of fake news detection. The complete table detailing these features is provided in the Appendix

\paragraph{Group 1: Fernandez and Devaraj Stylistic Features}

This study employed the collection of 34 linguistic attributes (referred to as ‘Linguistic Dimension’ and ‘Punctuation Cues’) that demonstrated the highest efficacy in classifying fake news, as determined through a sequence of tests outlined by \cite{Fernandez2019}. The inclusion of these features was motivated by \cite{Hoy2022} which used these features in providing preliminary evidence that stylistic features have the potential to be more generalisable than token-representations. But both studies relied on coarsely labelled datasets The two groups of features can be summarised as follows:

\begin{itemize}
    \item \textbf{Linguistic Dimensions:} Based on the Linguistic Dimensions of LIWC, this category aims to capture the complexity of news through inclusion of features such as word-per-sentence, average word size and type-token ratio (a measure of lexical variety) as well as the different types of words used such as the ratio of adjectives, nouns, verbs and named-entities.
    \item \textbf{Punctuation Cues:} Focuses solely on the different types of punctuation used relative to all punctuation in a given article.
\end{itemize}

As such linguistic features are a staple in NLP, the other groups of features described below also include similar groups of features. A comprehensive list of these features is provided in Table \ref{tab:fernandez}

\paragraph{Group 2: Abonizio Features}

This study leverages 21 features organised into three groups: complexity, stylometric and psychological. The inclusion of these features, similar to the previous group, is motivated by their use in another generalisability test on coarsely labelled datasets in \cite{Abonizio2020} Similar groups of features can also be found in \citep{Paschalides2019,Garg2022,Reis2019} thus motivating their inclusion in this study.

The ‘complexity’ and ‘stylometric’ features overlap with some of the features used in the previous ‘Fernandez’ feature-set however it should be noted that the Abonizio feature-set is not as granular. However, unlike the Fernandez feature-set, the Abonizio feature-set does extend to include a psychological category, capturing the sentiment analysis score of a given article.  A list of these features is provided in Table \ref{tab:abonizio} 

\paragraph{Group 3: Linguistic Inquiry Word Count (LIWC)}

As previously mentioned in Section 2, LIWC is a dictionary-based approach comprising of linguistic elements, punctuation characteristics as well as psycholinguistic features organised in a number of categories. These categories can be summarised as follows: 

\begin{itemize}
    \item \textbf{Summary Variables:} Aim to summarise the features from the below three categories and attempt to capture document complexity and psychological features.
    
    \item \textbf{Linguistic Dimensions:} Aims to capture different types of words such as pronouns, verbs and adjectives as well as words denoting grammatical person and numbers. 

    \item \textbf{Psychological Processes:} captures a number of psychological words relating to sentiment (such as ‘good’ and ‘bad’) as well as words relating to cognition (e.g. ‘know’ and ‘think’) and social processes (such as ‘love’ and ‘fight’).   

    \item \textbf{Expanded Dictionary:} captures a range of different words relating to a number of different topics such as culture and lifestyle as well as temporal words such as ‘when’, ‘now’ and ‘then’. 

\end{itemize}

The total number of features in this feature-set amount to 118. Similar to the previous feature groups, LIWC is used in a generalisability study by \cite{Perez-Rosas2018} which observed the performance of LIWC trained on the FakeNewsAMT dataset and tested Celebrity news datasets and vice versa. A number of other studies also leverage these features, thus further justifying their inclusion \citep{Ahmad2020,Spezzano2021,Shu2019BeyondCont} An exhaustive list of these features is provided in Table \ref{tab:LIWC}

\paragraph{Group 4: NELA Feature Extractor}

The NELA feature extractor is a tool hosted on GitHub designed by the authors of the NELA dataset which has been used throughout this study. This therefore motivated the inclusion of these features in this study. It includes a rich, hand-crafted feature-set of 91 features which can be summarised into the following categories:  

\begin{itemize}
    
    \item \textbf{Style:} Largely similar to those from the previous three studies, focussing on POS tags

    \item \textbf{Complexity:} Similar to the ‘Linguistic Dimensions’ and ‘Complexity’ categories of the Abonizio feature-set, this category of features aims to capture how complex an article is through analysing lexical diversity, reading-difficulty metrics and the average length of words and sentences

    \item \textbf{Bias:} Based on \cite{Recasens2013} work, this category of features aims to capture the subjectivity of the text by identifying the number of hedges, factives, assertives, implicatives, and opinion words.

    \item \textbf{Affect:} Relying on VADER sentiment analysis, this category aims to capture the emotion and sentiment of the text

    \item \textbf{Moral:} The objective of this feature category is to encompass the ethical content present in a text, and it is built upon the principles of Moral Foundation Theory (MFT) introduced by \cite{GRAHAM201355}.\cite{Lin2018} subsequently expanded upon this theory and developed a lexicon specifically designed for assessing the moral aspects of text. This feature group employs the lexicon established by \cite{Lin2018} to gauge the morality of the text under consideration.

    \item \textbf{Event:} Aims to capture words relating to dates, times and locations.

\end{itemize}

A full list of these features is provided in Table \ref{tab:NELA}

\paragraph{Group 5: Modified NELA Features}
Through the use of the NELA Feature Extractor, it was noted that a number of the features were either duplicated or returning zero values, in particular, when attempting to extract punctuation. As such, the NELA Feature Extractor was modified to remediate this and to include additional punctuation such as ‘\#’, ‘@’, ‘£’, ‘\$’, ‘\&’ and ‘\%’. The normalisation was also adjusted depending on the features. For example, rather than scaling the punctuation based on the word-count of an article, punctuation was scaled based on the total number of punctuation items. 

\subsubsection{Proposed `Social-Monetisation' Features}

As motivated in Section 2.5, a number of additional novel features, categorised as ‘social-monetisation’ features, were investigated to complement the groups of stylistic features selected for this study. These were: (i) the number of advertisements; (ii) the number of external links; (iii) the number of links to Facebook; and (iv) number of links to Twitter/X.  The number of ads was extracted through the use of EasyList, an open-source project that compiles a list of the most popular adblocking filters. Using this list enables searching the webpage’s LXML tree and counting the frequency of various ads. External links were identified through a combination of extracting ‘hrefs’ in the webpages and comparing their domains to the host domain using the ‘tldextract’ library. Links the domains of which did not match the host domain were used to calculate the frequency of external links. Links that pointed to Facebook and Twitter/X were each counted separately. 

\subsection{Machine Learning Algorithms}

As this study prioritises the exploration of stylistic features for generalisable fake news detection, less emphasis has been put on exploring the effect of different machine learning algorithms and their respective hyperparameters. However, for completeness and to offer an opportunity for comparison to the literature, a number of machine learning algorithms including Logistic Regression, SVM, Gradient Boosting, Decision Trees, Random Forest, and a feed-forward neural network (FFNN) are employed. Each of these algorithms was implemented using default hyperparameters in SKLearn. The exception to this was the neural network where default hyperparameters are not available and, as such, a shallow Sequential model was used with a single hidden layer of 10 neurons, a sigmoid activation layer compiled with binary cross-entropy loss and the Adam optimiser. To protect against overfitting, the EarlyStopping hyperparameter was set to stop training if the loss function did not improve by 0.01.

Additionally, a Long Short-Term Memory (LSTM) network was employed to process word embeddings generated by BERT and Word2Vec. LSTMs are well-suited for capturing sequential dependencies in embeddings, providing a deeper understanding of contextual relationships within the text. This approach was not applied to Bag-of-Words or TF-IDF features, as these methods represent text as sparse matrices, lacking sequential and contextual information, rendering them unsuitable for use with LSTMs. This distinction highlights the effort to align algorithms with feature sets that best leverage their strengths for fake news detection. The LSTM architecture used in this study consisted of an LSTM layer with 128 hidden units, designed to model complex temporal dependencies and contextual patterns in the data. A dropout rate of 40\% was applied to the LSTM's output to reduce the risk of overfitting. Finally, a fully connected layer was used to map the final hidden state to two output classes: fake and real news.

\subsection{Training/Testing Methodology}

The training and testing methodology in this study integrates K-fold cross-validation, external validation, and tailored evaluation techniques for large language models (LLMs) to assess performance and generalisability in fake news detection. For traditional machine learning algorithms and feature sets, K-fold cross-validation is employed on the NELA dataset, partitioning it into 10 equal folds. Models are trained on 9 folds and validated on the remaining fold across iterations, ensuring robust performance estimates with a fixed random state (set to 42). External  validation evaluates the generalisability of models by testing those trained on each NELA fold against 500 randomly sampled articles per class from the Facebook dataset, addressing class imbalance. Key evaluation metrics such as Accuracy, Precision, Recall, Specificity and F1-Score are used to assess model performance in distinguishing between true news' and fake news'. By evaluating the models using external validation after each fold.

For LLaMa zero-shot and few-shot models, a distinct methodology was adopted due to its pre-trained nature. In the zero-shot configuration, the model leveraged its existing knowledge without any task-specific training, relying entirely on structured prompts to classify text from both the entirety of the NELA dataset and 5 random balanced samples of 500 Facebook URLs dataset. In the few-shot configuration, the model was guided by labelled examples from the NELA dataset only to reflect the goals of this study to train on a coarsely labelled dataset but test on a manually labelled dataset. The fine-tuned LLaMa model followed a similar methodology to the above traditional methods.

In addition to the Mann-Whitney U-test,  Permutation Feature Importance (PFI) was used to pinpoint the stylistic and proposed social monetisation features that positively contributed to a more generalisable model. PFI works by shuffling a random feature, thereby disrupting the relationship between that feature and the target variable. By repeating this process for all features in the dataset and observing the effects on model performance, the method reveals how much the model depends on each feature

\section{Results}
This section outlines the results of the two experiments outlined in Section 4. The overarching objective of the experiments is to demonstrate whether models using different sets of stylistic and the proposed social-monetisation features are able to detect ‘real-world’ fake news (achieved by using the Facebook URLs dataset for evaluation) and comparing them to state-of-the-art approaches relying on token-representations and LLMs. 

\subsection{Experiment 1: Generalisability of Token-Representations \& LLMs}

Experiment 1 aimed to address RQ1 by examining the generalisation capabilities of token representations (BoW, TF-IDF, Word2Vec, and BERT, as detailed in Section 4.2.1) combined with different machine learning models, as well as evaluating the performance of the large language model LLaMa under zero-shot, few-shot, and fine-tuning configurations.

\subsubsection{Token-Representation Results}

\begin{table}[ht]
    \centering
    \caption{Token-Representations Baseline Results}
     \label{tab:tr-base}   
     \begin{adjustbox}{width=0.7\textwidth}
    \begin{tabular}{|l|l|l|l|l|l|l|} \hline  
      \textbf{Features}&\textbf{Model}&  \textbf{Acc.}&  \textbf{Prec.}&  \textbf{Rec.}&  \textbf{Spec.}& \textbf{F1}\\ \hline
      \multirow{6}{*}{BoW}&Logistic Regression&  0.98&  0.97&  0.99&  0.97& 0.98\\   
      & Decision Tree&  0.96&  0.95&  0.96&  0.95& 0.96\\   
      &SVM&  0.91&  0.85&  0.99&  0.84& 0.92\\   
      &Gradient Boosting&  0.97&  0.96&  0.99&  0.96& 0.97\\   
      &Random Forest&  0.99&  0.97&  1.00&  0.97& 0.99\\   
      &Neural Network&  0.99&  0.99&  0.98&  0.98& 0.99\\ \hline  
      \multirow{6}{*}{TF-IDF}&Logistic Regression
&  0.98&  0.96&  1.00&  0.96& 0.98\\ 
      &Decision Tree
&  0.95&  0.95&  0.96&  0.95& 0.95\\   
      &SVM
&  0.97&  0.94&  1.00&  0.94& 0.97\\   
      &Gradient Boosting
& 0.98& 0.96& 0.99& 0.97&0.98\\   
      &Random Forest
& 0.99& 0.97& 1.00& 0.97&0.99\\   
      &Neural Network& 0.99& 0.98& 1.00& 0.97&0.99\\ \hline  
      \multirow{6}{*}{Word2Vec}&Logistic Regression
& 0.89& 0.86& 0.93& 0.86&0.90\\   
      &Decision Tree
& 0.87& 0.85& 0.88& 0.88&0.86\\   
      &SVM
& 0.91& 0.86& 0.96& 0.85&0.91\\   
      &Gradient Boosting
& 0.95& 0.94& 0.97& 0.94&0.95\\   
      &Random Forest
& 0.95& 0.92& 0.98& 0.92&0.95 \\   
      &Neural Network& 0.88& 0.91& 0.86& 0.83&0.88 \\
      &LSTM& 0.88& 0.91& 0.84& 0.91&0.87 \\ \hline  
      \multirow{6}{*}{BERT}&Logistic Regression
& 0.88& 0.86& 0.90& 0.86&0.88 \\   
      &Decision Tree
& 0.81& 0.82& 0.80& 0.82&0.81 \\   
      &SVM
& 0.90& 0.89& 0.92& 0.89&0.90 \\   
      &Gradient Boosting
& 0.84& 0.83& 0.85& 0.83&0.84 \\   
      &Random Forest
& 0.84& 0.82& 0.86& 0.82&0.84 \\   
      &Neural Network& 0.85& 0.89& 0.81& 0.83&0.84 \\ 
      &LSTM& 0.99& 0.99& 0.99& 0.99&0.99\\
      &Fine-Tuned BERT& 1.0& 1.0& 1.0& 1.0&1.0\\ \hline 
    \end{tabular}
    \end{adjustbox}

\end{table}

As seen in Table \ref{tab:tr-base}, under K-fold test conditions, token-representations trained and tested on the NELA dataset exhibit high performance across several models, achieving a mean accuracy of 0.93, with a range between 0.79 and 1.0. These results align with those commonly reported in the literature and validate the effectiveness of the feature extraction and modeling methods when applied in controlled conditions.

Among the tested approaches, Fine-Tuned BERT and LSTMs demonstrated the highest performance, achieving near-perfect metrics. Fine-Tuned BERT reached 1.0 accuracy, precision, recall, specificity, and F1 score, while LSTMs trained on BERT embeddings achieved 0.99 across all metrics. However, such results may reflect the model's ability to memorise patterns specific to the training data rather than a genuine ability to generalise beyond the dataset.

Traditional feature extraction techniques such as BoW and TF-IDF also achieved strong results, with accuracies up to 0.99. Their reliance on high-frequency or significant terms may have contributed to their robust performance within the dataset. This aligns with our observations in Section 2.4 regarding biases within commonly used fake news datasets, where certain terms may strongly correlate with specific classes. Specifically, Logistic Regression outperformed SVM when using the BoW representation, likely due to its ability to leverage the sparse, linearly separable nature of BoW features. In contrast, the SVM with an RBF kernel may not have been optimally suited for this representation, as the kernel is designed for capturing non-linear relationships, which may not align with BoW’s characteristics \citep{Colas2007}. This mismatch could partly explain the relatively poorer performance of SVM in this context.

In contrast to BoW and TF-IDF, Word2Vec and standard BERT-base embeddings performed slightly worse, with accuracies ranging from 0.81 to 0.95. These models may have struggled with the added noise in the textual data or the more complex representations introduced by their embeddings. In the case of Word2Vec, despite using a relevant model pre-trained on Google News, the performance may have been hindered by out-of-vocabulary (OOV) words. Unlike BoW and TF-IDF, which construct their vocabularies directly from the dataset and therefore capture all words within that domain, Word2Vec relies on a fixed vocabulary from its pre-training corpus. Fake news datasets often contain domain-specific terms, slang, or creative language use that may not be present in the pre-trained Word2Vec vocabulary. As a result, OOV words are either ignored or mapped to suboptimal representations, leading to a potential loss of crucial information. This limitation reduces the model’s ability to capture dataset-specific keywords and patterns, whereas simpler methods like BoW and TF-IDF, by leveraging dataset-dependent vocabularies, retain the ability to represent all terms present in the text.

While these results highlight the apparent effectiveness of both traditional and deep learning approaches in controlled settings, they must be interpreted cautiously. The high performance observed here may not translate to real-world applications or unseen datasets, as will be discussed in the cross-dataset results.

\begin{table}[ht]
    \centering
    \caption{Token-Representations Cross-Dataset Results}
     \label{tab:tr-xd}   
     \begin{adjustbox}{width=0.7\textwidth}
   \begin{tabular}{|l|l|l|l|l|l|l|} \hline  
      \textbf{Features}&\textbf{Model}&  \textbf{Acc.}&  \textbf{Prec.}&  \textbf{Rec.}&  \textbf{Spec.}& \textbf{F1}\\ \hline
      \multirow{6}{*}{BoW}&Logistic Regression&  0.66&  0.68&  0.60&  0.72& 0.64
\\   
      & Decision Tree&  0.64&  0.66&  0.57&  0.71& 0.61
\\   
      &SVM&  0.61&  0.58&  0.78&  0.44& 0.67
\\   
      &Gradient Boosting&  0.63&  0.60&  0.76&  0.50& 0.67
\\   
      &Random Forest&  0.61&  0.57&  0.88&  0.34& 0.69
\\   
      &Neural Network&  0.68&  0.72&  0.58&  0.78& 0.65
\\ \hline  
      \multirow{6}{*}{TF-IDF}&Logistic Regression
&  0.68&  0.68&  0.71&  0.66& 0.69
\\ 
      &Decision Tree
&  0.64&  0.67&  0.58&  0.71& 0.62
\\   
      &SVM
&  0.68&  0.66&  0.74&  0.62& 0.70
\\   
      &Gradient Boosting
& 0.64& 0.62& 0.76& 0.53&0.68
\\   
      &Random Forest
& 0.64& 0.60& 0.86& 0.42&0.70
\\   
      &Neural Network& 0.70& 0.74& 0.63& 0.78&0.68
\\ \hline  
      \multirow{6}{*}{Word2Vec}&Logistic Regression
& 0.67& 0.70& 0.57& 0.76&0.63
\\   
      &Decision Tree
& 0.60& 0.62& 0.56& 0.66&0.59
\\   
      &SVM
& 0.65& 0.68& 0.57& 0.73&0.62
\\   
      &Gradient Boosting
& 0.65& 0.67& 0.58& 0.72&0.62
\\   
      &Random Forest
& 0.65& 0.66& 0.62& 0.67&0.64
\\   
      &Neural Network& 0.66& 0.70& 0.55& 0.77&0.61\\ 
      &LSTM& 0.62& 0.64& 0.54& 0.7&0.59\\\hline  
      \multirow{6}{*}{BERT}&Logistic Regression
& 0.65& 0.67& 0.60& 0.70&0.63
\\   
      &Decision Tree
& 0.60& 0.62& 0.53& 0.67&0.57
\\   
      &SVM
& 0.66& 0.69& 0.59& 0.74&0.64
\\   
      &Gradient Boosting
& 0.63& 0.64& 0.61& 0.66&0.62
\\   
      &Random Forest
& 0.62& 0.62& 0.62& 0.62&0.62
\\   
      &Neural Network& 0.66& 0.69& 0.56& 0.75&0.62\\ 
      &LSTM& 0.68& 0.75& 0.55& 0.81&0.63\\
      &Fine-Tuned BERT& 0.68& 0.75& 0.53& 0.82&0.62\\
      \hline 
    \end{tabular}
    \end{adjustbox}

\end{table}

The results from the cross-dataset testing (Table \ref{tab:tr-xd}) reveal a significant reduction in performance across all models and feature sets compared to the K-fold test conditions discussed in the previous section. While token-representation models achieved high accuracy within the NELA dataset (Section \ref{tab:tr-base}), their ability to generalise across datasets is markedly lower. On average, models experienced a 0.28 drop in accuracy, underscoring the challenge of applying these approaches to unseen data.

Among the tested feature sets, Fine-Tuned BERT and LSTM models, which previously excelled in the controlled K-fold setting, demonstrated similar vulnerabilities to generalisation issues. Fine-Tuned BERT achieved an accuracy of 0.68, precision of 0.75, recall of 0.53, specificity of 0.82, and an F1 score of 0.62. While these metrics are comparable to other models in cross-dataset conditions, they represent a stark decline from its perfect performance in K-fold testing. Similarly, the LSTM trained on BERT embeddings exhibited similar limitations, achieving a slightly lower accuracy of 0.62 and an imbalanced recall and specificity, suggesting a difficulty in maintaining consistency across datasets.

Traditional feature sets such as BoW and TF-IDF also showed substantial performance degradation, with accuracy ranging between 0.61 and 0.7. While Logistic Regression and SVM models trained on these features demonstrated relatively balanced recall and specificity, their overall drop in accuracy highlights the limitations of token-representation methods when applied to different datasets. Word2Vec and BERT, which already underperformed compared to BoW and TF-IDF in K-fold testing, fared similarly poorly under cross-dataset conditions, suggesting that they may be more sensitive to noise in the training data.

Neural networks trained on TF-IDF features achieved the highest accuracy under cross-dataset conditions (0.7), but their poor recall indicates that they struggled to correctly identify the positive (true news) class. This trend, observed across other neural network models, suggests potential overfitting to the NELA dataset, even with the inclusion of the early-stopping hyperparameter. In contrast, SVM and Logistic Regression models trained on TF-IDF features exhibited more balanced recall and specificity, making them arguably better candidates for generalisation despite their slightly lower overall accuracy.

These findings underscore a key limitation of token-representation-based models: while they can achieve high performance in controlled conditions, their generalisability to unseen datasets remains limited. This highlights the need for exploring alternative feature sets beyond the text, such as stylistic or social-monetisation features, which may offer more robust solutions for addressing dataset variability and bias. Additionally, the large imbalances observed between precision, recall, and specificity across models suggest the influence of underlying differences in how these methods respond to dataset-specific patterns, further complicating generalisability.

\subsection{LLaMa Results}

This section evaluates the performance of LLaMa 3.2-1B in detecting fake news, focusing on its zero-shot, few-shot, and fine-tuned configurations. Unlike traditional algorithms relying on static token representations, LLaMa leverages its pre-trained, context-aware transformer architecture to adapt dynamically to the task.

\begin{table}[!h]
    \centering
    \resizebox{0.6\textwidth}{!}{%
    \renewcommand{\arraystretch}{1.2}
    \begin{tabular}{|l|c|c|c|c|c|}
        \hline
        \textbf{Config.} & \textbf{Acc.} & \textbf{Prec.} & \textbf{Rec.} & \textbf{Spec.} & \textbf{F1} \\ \hline
        Zero-Shot & 0.53 & 0.55 & 0.39 & 0.68 & 0.46 \\ 
        Few-Shot  & 0.60 & 0.65 & 0.42 & 0.77 & 0.51 \\ 
        Fine-Tuned & 1 & 1 & 1 & 1.00 & 1 \\ \hline
    \end{tabular}}
    \caption{Performance of LLaMa configurations on the NELA dataset.}
    \label{tab:llama_nela_results}
\end{table}

The results presented in Table \ref{tab:llama_nela_results} reveal the performance of LLaMa 3.2-1B under zero-shot, few-shot, and fine-tuned configurations on the NELA dataset.

In the zero-shot configuration, LLaMa achieved an accuracy of only 0.53, with precision at 0.55 and a low recall of 0.39. The specificity was 0.68, and the F1-score stood at 0.46. These metrics are barely above random chance, indicating that the model struggled to effectively detect fake news without any task-specific training. Despite large language models often performing well in zero-shot settings for other tasks, LLaMa's performance here suggests that its pre-trained knowledge does not generalise well to the nuances of fake news detection.

The few-shot configuration showed a modest improvement, with accuracy increasing to 0.6, precision to 0.65, and recall to 0.42. Specificity improved to 0.77, and the F1-score to 0.51. While these results are slightly better than the zero-shot configuration, they remain unsatisfactory for practical applications. The limited enhancement implies that providing a small number of labeled examples was insufficient for the model to grasp the complex patterns associated with fake news, which often involve subtle linguistic cues and context-dependent nuances.

In stark contrast, the fine-tuned configuration achieved perfect scores across all metrics, with accuracy, precision, recall, specificity, and F1-score all at 1.00. While this suggests that the model can perform exceptionally well when extensively trained on the task-specific data, such flawless performance is unusual and may indicate overfitting to the NELA dataset. Overfitting reduces the model's ability to generalise to new, unseen data, limiting its practical utility in real-world scenarios where fake news can vary widely in form and content.

These findings underscore a critical limitation of large language models like LLaMa in the context of fake news detection. Despite their strong performance in zero-shot and few-shot settings on more general language tasks, these models do not perform well on the task of fake news detection without substantial task-specific training.

\begin{table}[!h]
    \centering
    \resizebox{0.8\textwidth}{!}{%
    \renewcommand{\arraystretch}{1.2}
    \begin{tabular}{|l|c|c|c|c|c|}
        \hline
        \textbf{Configuration} & \textbf{Accuracy} & \textbf{Precision} & \textbf{Recall} & \textbf{Specificity} & \textbf{F1} \\ \hline
        Zero-Shot & 0.50 & 0.50 & 0.40 & 0.61 & 0.44 \\ 
        Few-Shot  & 0.55 & 0.57 & 0.37 & 0.72 & 0.45 \\ 
        Fine-Tuned & 0.71 & 0.90 & 0.47 & 0.95 & 0.62 \\ \hline
    \end{tabular}}
    \caption{Performance of LLaMa configurations on the Facebook URLs dataset.}
    \label{tab:llama_facebook_results}
\end{table}

This issue is demonstrated further in regards to the Facebook URLs dataset, where Table \ref{tab:llama_facebook_results} highlights the cross-dataset performance of LLaMa in these three configurations. These findings provide insights into the model's ability to generalise when tested on unseen, manually labelled data.

In the zero-shot configuration, LLaMa achieves an accuracy of 0.5, with precision and recall at 0.5 and 0.4, respectively. The specificity is moderately better at 0.61, and the F1-score stands at 0.44. These results indicate that the model performs at near-random levels when applied to the external dataset without any task-specific training. While large language models are often effective in zero-shot scenarios for general tasks, LLaMa's performance here underscores the difficulty of adapting pre-trained knowledge to the domain-specific challenges of fake news detection, particularly when confronted with nuanced and diverse real-world data.

The few-shot configuration using examples from the NELA dataset shows a marginal improvement over zero-shot performance. Accuracy increases to 0.55, precision rises to 0.57, and specificity improves to 0.72. However, recall remains low at 0.37, and the F1-score only improves slightly to 0.45. These results suggest that while a small number of labeled examples from the training dataset provided some task-specific guidance, they were insufficient for LLaMa to effectively generalise to the Facebook dataset. This limited improvement highlights the challenges of adapting models trained on coarsely labelled datasets, such as NELA, to manually curated datasets with more nuanced distinctions.

In the fine-tuned configuration of LLaMA trained on the NELA dataset, the model demonstrates a notable improvement, achieving an accuracy of 0.71, precision of 0.9 for identifying fake news (the negative class), and specificity of 0.95. However, recall remains relatively low at 0.47, resulting in an F1-score of 0.62. While the accuracy is higher compared to token-representations, the high specificity suggests the model is effective at identifying fake news, but the low recall coupled with high precision indicates it is conservative in predicting true news, potentially overlooking many true news instances. Similar imbalanced metrics can be observed in token-based approaches, suggesting that both LLMs and token-representations face challenges in achieving balanced performance across classes, likely due to dataset-specific biases and the inherent difficulty of the classification task.

\subsection{Experiment 2: Generalisability of Stylistic \& Social Monetisation \\Features}

The second experiment targeted RQ2 and RQ3 and aimed to determine whether the stylistic features suggested in the literature and the social-monetisation features introduced in this study are more generalisable than the token-level representations tested in Section 5.1. As detailed in Section 4.3.1, the following groups of stylistic features were evaluated: Fernandez; Abonizio; LIWC; NELA; and modified NELA. Each of these groups was tested with and without the proposed social monetisation features identified in Section 4.3.2. A K-fold test was first performed with the same splits as in the first experiment, using the NELA dataset to provide a baseline for comparison and the Facebook dataset to perform a cross-dataset test for each model trained in each fold.

\begin{table}[ht]
    \begin{adjustwidth}{-.5in}{-.5in} 
    \centering
    \caption{Stylistic Features \& S-M Features Baseline Results}
    \label{tab:style-base}
    \begin{adjustbox}{width=1.15\textwidth}
    \begin{tabular}{|l|l|c|c|c|c|c|c|c|c|c|c|l|}
        \hline
        \multirow{2}{*}{\textbf{Feature-Set}} & \multirow{2}{*}{\textbf{Model}} & \multicolumn{5}{c|}{\textbf{Without proposed S-M Features}} & \multicolumn{5}{c|}{\textbf{With proposed S-M Features}} & \multirow{2}{*}{\textbf{p-value}}\\ \cline{3-12} 
         &  & \textbf{Acc.} & \textbf{Prec.} & \textbf{Rec.} & \textbf{Spec.} & \textbf{F1} & \textbf{Acc.} & \textbf{Prec.} & \textbf{Rec.} & \textbf{Spec.} & \textbf{F1}   &\\ \hline
        \multirow{6}{*}{Fernandez} & Logistic Regression & 0.83& 0.80& 0.89& 0.75& 0.84& 0.84& 0.81& 0.90& 0.78& 0.85 &{\cellcolor[gray]{.8}} \textless0.001\\ 
         & Decision Tree & 0.88& 0.88& 0.88& 0.87& 0.88& 0.93& 0.93& 0.93& 0.93& 0.93 &{\cellcolor[gray]{.8}} \textless0.001\\ 
         & SVM & 0.90& 0.86& 0.95& 0.84& 0.91& 0.92& 0.89& 0.96& 0.87& 0.93 &{\cellcolor[gray]{.8}} \textless0.001\\ 
         & Gradient Boosting & 0.89& 0.87& 0.93& 0.85& 0.90& 0.93& 0.92& 0.96& 0.91& 0.94&{\cellcolor[gray]{.8}} \textless0.001\\ 
         & Random Forest & 0.93& 0.91& 0.96& 0.90& 0.94& 0.96& 0.95& 0.98& 0.95& 0.97&{\cellcolor[gray]{.8}} \textless0.001\\ 
         & Neural Network & 0.89& 0.87& 0.92& 0.85& 0.90& 0.93& 0.92& 0.95& 0.91& 0.93&{\cellcolor[gray]{.8}} \textless0.001\\ \hline
        \multirow{6}{*}{Abonizio} & Logistic Regression & 0.78& 0.77& 0.82& 0.73& 0.79& 0.78& 0.77& 0.82& 0.74& 0.79 &{\cellcolor[gray]{.8}} 0.5678\\ 
         & Decision Tree & 0.85& 0.86& 0.86& 0.85& 0.86& 0.92& 0.92& 0.92& 0.91& 0.92 &{\cellcolor[gray]{.8}} \textless0.001\\ 
         & SVM & 0.91& 0.89& 0.94& 0.88& 0.91& 0.94& 0.93& 0.97& 0.92& 0.95&{\cellcolor[gray]{.8}} \textless0.001\\ 
         & Gradient Boosting & 0.86& 0.85& 0.90& 0.82& 0.87& 0.93& 0.91& 0.96& 0.90& 0.93&{\cellcolor[gray]{.8}} \textless0.001\\ 
         & Random Forest & 0.93& 0.92& 0.95& 0.91& 0.94& 0.98& 0.97& 0.99& 0.97& 0.98&{\cellcolor[gray]{.8}} \textless0.001\\ 
         & Neural Network & 0.90& 0.90& 0.92& 0.88& 0.91& 0.94& 0.94& 0.96& 0.93& 0.95&{\cellcolor[gray]{.8}} \textless0.001\\ \hline
        \multirow{6}{*}{LIWC}& Logistic Regression & 0.91& 0.90& 0.92& 0.89& 0.91& 0.91& 0.91& 0.92& 0.90& 0.92 & {\cellcolor[gray]{.8}} 0.2017\\ 
         & Decision Tree & 0.88& 0.88& 0.88& 0.88& 0.88& 0.90& 0.91& 0.90& 0.91& 0.91&{\cellcolor[gray]{.8}} \textless0.001\\ 
         & SVM & 0.97& 0.96& 0.98& 0.95& 0.97& 0.97& 0.96& 0.98& 0.96& 0.97&{\cellcolor[gray]{.8}} 0.03\\ 
         & Gradient Boosting & 0.94& 0.92& 0.96& 0.91& 0.94& 0.95& 0.94& 0.97& 0.93& 0.95&{\cellcolor[gray]{.8}} \textless0.001\\ 
         & Random Forest & 0.95& 0.93& 0.98& 0.92& 0.95& 0.96& 0.94& 0.99& 0.94& 0.96&{\cellcolor[gray]{.8}} \textless0.001\\ 
         & Neural Network & 0.95& 0.95& 0.96& 0.94& 0.95& 0.96& 0.95& 0.96& 0.95& 0.96&{\cellcolor[gray]{.8}} 0.03\\ \hline
        \multirow{6}{*}{\shortstack[l]{NELA Feature\\Extractor}}& Logistic Regression & 0.85& 0.85& 0.88& 0.83& 0.86& 0.86& 0.85& 0.88& 0.84& 0.87& {\cellcolor[gray]{.8}} 0.023\\ 
         & Decision Tree & 0.85& 0.86& 0.85& 0.85& 0.86& 0.89& 0.89& 0.89& 0.89& 0.89&{\cellcolor[gray]{.8}} \textless0.001\\ 
         & SVM & 0.94& 0.92& 0.96& 0.91& 0.94& 0.95& 0.93& 0.97& 0.92& 0.95&{\cellcolor[gray]{.8}} \textless0.001\\ 
         & Gradient Boosting & 0.90& 0.88& 0.93& 0.86& 0.91& 0.93& 0.91& 0.96& 0.89& 0.93&{\cellcolor[gray]{.8}} \textless0.001\\ 
         & Random Forest & 0.93& 0.90& 0.97& 0.88& 0.93& 0.95& 0.93& 0.98& 0.92& 0.95&{\cellcolor[gray]{.8}} \textless0.001\\ 
         & Neural Network & 0.92& 0.91& 0.93& 0.91& 0.92& 0.94& 0.94& 0.95& 0.93& 0.94&{\cellcolor[gray]{.8}} \textless0.001\\ \hline
        \multirow{6}{*}{\shortstack[l]{Modified NELA \\Feature Extractor}}& Logistic Regression & 0.87& 0.87& 0.89& 0.86& 0.88& 0.88& 0.88& 0.89& 0.86& 0.88&{\cellcolor[gray]{.8}} 0.023\\ 
         & Decision Tree & 0.88& 0.89& 0.88& 0.88& 0.88& 0.90& 0.91& 0.91& 0.90& 0.91&{\cellcolor[gray]{.8}} \textless0.001\\ 
         & SVM & 0.96& 0.95& 0.98& 0.94& 0.96& 0.97& 0.96& 0.98& 0.95& 0.97&{\cellcolor[gray]{.8}} \textless0.001\\ 
         & Gradient Boosting & 0.92& 0.90& 0.94& 0.89& 0.92& 0.94& 0.93& 0.96& 0.92& 0.94&{\cellcolor[gray]{.8}} \textless0.001\\ 
         & Random Forest & 0.95& 0.93& 0.97& 0.92& 0.95& 0.97& 0.95& 0.98& 0.94& 0.97&{\cellcolor[gray]{.8}} \textless0.001\\ 
         & Neural Network & 0.94& 0.94& 0.95& 0.94& 0.95& 0.95& 0.95& 0.96& 0.95& 0.96&{\cellcolor[gray]{.8}} \textless0.001\\ \hline
    \end{tabular}
    \end{adjustbox}
    \end{adjustwidth}
\end{table}

As can be seen from Table \ref{tab:style-base}, in K-fold cross-validation test conditions the selected stylistic features performed comparably to the token-representations (see Table \ref{tab:tr-base}). Across the different groups of stylistic features and machine learning algorithms, the mean accuracy was 90\% with a range between 78\% and 98\%. From this test, it can be seen that that Logistic Regression models using the Fernandez and Abonizio feature-sets excluding the proposed social monetisation features perform least well with the Random Forest trained on the Abonizio feature-group performing the best. Furthermore, with the inclusion of the proposed features, each feature group performed marginally better than without them. This is supported by the results of the Mann-Whitney U-test, whereby the p-values across the majority of models indicate a statistically significant increase in mean accuracy in models using the proposed features compared to those that exclude these features. Additionally, the modified NELA features also performed slightly better compared to the original NELA feature-set both with and without the proposed features. These results provide support for the use of the proposed features as well as the modification of the original NELA feature-set. 

\begin{table}[ht]
    \begin{adjustwidth}{-.5in}{-.5in} 
    \centering
    \caption{Stylistic Features \& S-M Features Cross-Dataset Results}
    \label{tab:style-xd}
    \begin{adjustbox}{width=1.15\textwidth}
    \begin{tabular}{|l|l|c|c|c|c|c|c|c|c|c|c|l|}
        \hline
        \multirow{2}{*}{\textbf{Feature-Set}} & \multirow{2}{*}{\textbf{Model}} & \multicolumn{5}{c|}{\textbf{Without proposed S-M Features}} & \multicolumn{5}{c|}{\textbf{With proposed S-M Features}} & \multirow{2}{*}{\textbf{p-value}}\\ \cline{3-12} 
         &  & \textbf{Acc.} & \textbf{Prec.} & \textbf{Rec.} & \textbf{Spec.} & \textbf{F1} & \textbf{Acc.} & \textbf{Prec.} & \textbf{Rec.} & \textbf{Spec.} & \textbf{F1}   &\\ \hline
        \multirow{6}{*}{Fernandez} 
& Logistic Regression & 0.66& 0.66& 0.67& 0.65& 0.67& 0.68& 0.67& 0.70& 0.66& 0.68&{\cellcolor[gray]{.8}} \textless0.001\\     
& Decision Tree & 0.66& 0.65& 0.67& 0.65& 0.66& 0.68& 0.69& 0.64& 0.71& 0.66&{\cellcolor[gray]{.8}} \textless0.001\\ 
& SVM & 0.68& 0.66& 0.71& 0.64& 0.69& 0.69& 0.69& 0.71& 0.68& 0.70&{\cellcolor[gray]{.8}} \textless0.001\\ 
& Gradient Boosting & 0.73& 0.73& 0.73& 0.73& 0.73& 0.74& 0.75& 0.72& 0.76& 0.74&{\cellcolor[gray]{.8}} \textless0.001\\ 
& Random Forest & 0.73& 0.76& 0.67& 0.79& 0.71& 0.74& 0.76& 0.70& 0.78& 0.72&{\cellcolor[gray]{.8}} \textless0.001\\ 
& Neural Network & 0.70& 0.70& 0.69& 0.71& 0.69& 0.71& 0.71& 0.69& 0.72& 0.70&{\cellcolor[gray]{.8}} 0.4446\\ \hline
        \multirow{6}{*}{Abonizio} 
& Logistic Regression & 0.59& 0.59& 0.62& 0.56& 0.60& 0.62& 0.61& 0.66& 0.57& 0.63&{\cellcolor[gray]{.8}} \textless0.001\\ 
& Decision Tree & 0.57& 0.57& 0.56& 0.58& 0.57& 0.60& 0.60& 0.60& 0.59& 0.60&{\cellcolor[gray]{.8}} \textless0.001\\ 
& SVM & 0.63& 0.63& 0.62& 0.64& 0.62& 0.65& 0.66& 0.64& 0.67& 0.65&{\cellcolor[gray]{.8}} \textless0.001\\ 
& Gradient Boosting & 0.61& 0.60& 0.64& 0.58& 0.62& 0.65& 0.65& 0.67& 0.63& 0.66&{\cellcolor[gray]{.8}} \textless0.001\\ 
& Random Forest & 0.64& 0.65& 0.61& 0.67& 0.63& 0.67& 0.68& 0.67& 0.68& 0.67&{\cellcolor[gray]{.8}} \textless0.001\\ 
& Neural Network & 0.60& 0.60& 0.59& 0.61& 0.60& 0.64& 0.64& 0.61& 0.66& 0.63&{\cellcolor[gray]{.8}} \textless0.001\\ \hline
        \multirow{6}{*}{LIWC}
& Logistic Regression & 0.69& 0.71& 0.65& 0.73& 0.68& 0.67& 0.69& 0.62& 0.72& 0.65&{\cellcolor[gray]{.8}}  N/A\\ 
& Decision Tree & 0.62& 0.62& 0.63& 0.61& 0.62& 0.62& 0.62& 0.63& 0.61& 0.62&{\cellcolor[gray]{.8}} 0.3564\\ 
& SVM & 0.67& 0.68& 0.65& 0.70& 0.66& 0.68& 0.68& 0.67& 0.68& 0.67&{\cellcolor[gray]{.8}} 0.02\\ 
& Gradient Boosting & 0.69& 0.70& 0.65& 0.72& 0.68& 0.74& 0.74& 0.72& 0.75& 0.73&{\cellcolor[gray]{.8}} \textless0.001\\ 
& Random Forest & 0.67& 0.67& 0.69& 0.66& 0.68& 0.69& 0.69& 0.68& 0.69& 0.69&{\cellcolor[gray]{.8}} \textless0.01\\ 
& Neural Network & 0.68& 0.69& 0.64& 0.71& 0.66& 0.69& 0.71& 0.63& 0.74& 0.67&{\cellcolor[gray]{.8}} 0.0319\\ \hline
        \multirow{6}{*}{\shortstack[l]{NELA Feature\\Extractor}}
& Logistic Regression & 0.65& 0.65& 0.63& 0.67& 0.64& 0.68& 0.69& 0.65& 0.71& 0.67& {\cellcolor[gray]{.8}} \textless0.001\\ 
& Decision Tree & 0.61& 0.61& 0.58& 0.63& 0.60& 0.62& 0.63& 0.61& 0.64& 0.62&{\cellcolor[gray]{.8}} 0.0258\\ 
& SVM & 0.63& 0.63& 0.61& 0.65& 0.62& 0.65& 0.66& 0.61& 0.69& 0.63&{\cellcolor[gray]{.8}} \textless0.001\\ 
& Gradient Boosting & 0.69& 0.72& 0.64& 0.75& 0.68& 0.70& 0.71& 0.67& 0.73& 0.69&{\cellcolor[gray]{.8}} 0.1312\\ 
& Random Forest & 0.66& 0.68& 0.62& 0.70& 0.65& 0.68& 0.69& 0.64& 0.72& 0.66&{\cellcolor[gray]{.8}} \textless0.001\\ 
& Neural Network & 0.63& 0.64& 0.60& 0.67& 0.62& 0.65& 0.67& 0.61& 0.70& 0.64&{\cellcolor[gray]{.8}} \textless0.001\\ \hline
        \multirow{6}{*}{\shortstack[l]{Modified NELA\\Feature Extractor}}
& Logistic Regression & 0.67& 0.68& 0.63& 0.70& 0.65& 0.71& 0.73& 0.67& 0.75& 0.70&{\cellcolor[gray]{.8}} \textless0.001\\ 
& Decision Tree & 0.64& 0.64& 0.63& 0.65& 0.64& 0.66& 0.66& 0.65& 0.66& 0.66&{\cellcolor[gray]{.8}} \textless0.001\\ 
& SVM & 0.66& 0.66& 0.65& 0.67& 0.66& 0.69& 0.70& 0.64& 0.73& 0.67&{\cellcolor[gray]{.8}} \textless0.001\\ 
& Gradient Boosting & 0.70& 0.70& 0.70& 0.71& 0.70& 0.75& 0.76& 0.72& 0.77& 0.74&{\cellcolor[gray]{.8}} \textless0.001\\ 
& Random Forest & 0.69& 0.70& 0.67& 0.72& 0.69& 0.75& 0.76& 0.73& 0.77& 0.74&{\cellcolor[gray]{.8}} \textless0.001\\ 
         & Neural Network & 0.64& 0.65& 0.62& 0.67& 0.63& 0.69& 0.70& 0.64& 0.73& 0.67&{\cellcolor[gray]{.8}} \textless0.001\\ \hline
    \end{tabular}
    \end{adjustbox}
    \end{adjustwidth}
\end{table}

In terms of generalisability, Table \ref{tab:style-xd} shows most models utilising stylistic features showed slightly better performance compared to LLaMa and models using token representations in the cross-dataset test, with an average drop in accuracy between the baseline and cross-dataset test of 24\%. While not all models trained on stylistic features outperformed token-based models or LLaMa, they consistently demonstrated a more balanced performance in recall and specificity, effectively addressing key limitations observed in models relying on token representations or LLaMa. Some models also demonstrated superior cross-dataset accuracy compared to the best performing models in the previous experiment, the fine-tuned LLaMa model, as well as the best performing model trained on token-representations (the Neural Network trained on TF-IDF features). Among those models trained without the proposed social-monetisation features, both Gradient Boosting and Random Forest algorithms trained on the Fernandez feature-set exhibited higher mean accuracy than LLaMa or any models trained using token-representations. The number of models displaying higher mean accuracy than those trained on token representations/LLaMa increased after integrating the proposed social-monetisation features, suggesting that the proposed features are relevant to producing more generalisable fake news detection models. Models incorporating these features and outperforming both token-representations and LLaMa include Gradient Boosting, and Random Forest trained on the modified NELA feature-set; Gradient Boosting trained on LIWC; and Gradient Boosting and Random Forest trained on the Fernandez feature-set. The Mann-Whitney U-test provides statistical support that the proposed features have a significant contribution in terms of improving the generalisability of these models. 

\subsection{Analysis with Permutation Feature Importance}

Experiment 2 presented evidence that the proposed social monetisation features contribute to producing more generalisable models. Permutation Feature Importance (PFI) analysis will further assess the impact of these features on the generalisability of the model. To prevent redundancy, the most successful model was selected for this analysis, which was Gradient Boosting trained on the modified NELA feature set, due to its higher mean accuracy (75\%) in cross-dataset conditions compared to other models. Although Random Forest trained on the same feature set demonstrates similar superior mean accuracy, Gradient Boosting was preferred due to its better performance across the other feature sets when compared to Random Forest. PFI was implemented by training the model on the NELA dataset and calculating the feature importance on both an unseen portion of the NELA dataset and a random balanced sample of the Facebook URLs dataset. This allows us to observe the features that are relevant to both models, and therefore what features can be considered the most generalisable between the coarsely labelled NELA dataset and the manually-labelled Facebook URLs dataset. 

Figures \ref{fig:baseline-pfi} and \ref{fig:external-pfi} display the feature importance plots for the Gradient Boosting models used in fake news detection, highlighting features that contribute meaningfully to model predictions. Due to the nature of the Gradient Boosting algorithm, certain features with `zero' importance were excluded from the plots. This exclusion likely results from the algorithm's tendency to select only one feature among highly correlated ones, thereby focusing on features with distinct positive or negative impacts on model performance.

\begin{table}[!h]
    \centering
    \resizebox{0.8\textwidth}{!}{ 
    \renewcommand{\arraystretch}{0.9}
    \small
    \begin{tabular}{|l|p{8cm}|}
        \hline
        \textbf{Feature} & \textbf{Description} \\ \hline
         ads              & Number of advertisements \\ 
         all\_caps        & Words written entirely in uppercase \\ 
         ampersand        & Frequency of ampersand characters (\&)\\ 
         at               & Frequency of the ``at" symbol (@) \\ 
         avg\_wordlen     & Average length of words \\ 
         CD               & Cardinal numbers  \\ 
         coleman\_liau\_index & Readability metric indicating grade level \\ 
         dollar           & Dollar signs (\$) \\ 
         exclamation      & Exclamation marks (!) \\ 
         ext\_total       & Total number of external links \\ 
         fb               & Presence of Facebook-related content \\ 
         FW               & Foreign words \\ 
         IngroupVirtue    & Words conveying positive group associations \\ 
         JJR              & Comparative adjectives (e.g., better) \\ 
         JJS              & Superlative adjectives (e.g., best) \\ 
         NNP              & Singular proper nouns \\ 
         NNPS             & Plural proper nouns \\ 
         percentage       & Percentage signs (\%) \\ 
         POS              & Part-of-speech tags \\ 
         PurityVice       & Words indicating impurity or moral vice \\ 
         question         & Question marks (?) \\ 
         RP               & Particles\\ 
         single\_quote    & Single quotation marks (') \\ 
         stops            & Stop words\\ 
         TO               & Infinitive marker ``to" \\ 
         ttr              & Type-token ratio (lexical diversity) \\ 
         twit             & Presence of Twitter-related content \\ 
         vadneu           & Neutral valence in sentiment analysis \\ 
         vadpos           & Positive valence in sentiment analysis \\ 
         VB               & Base form verbs \\ 
         VBN              & Past participle verbs \\ 
         WDT              & Wh-determiners (e.g., which) \\ 
         word\_count      & Total number of words\\ \hline
    \end{tabular}}
    \caption{Relevant features to both datasets}
    \label{tab:positive-feature-rankings}
\end{table}

In examining these plots, we can identify 33 features (Table \ref{tab:positive-feature-rankings}) that hold relevance across both datasets, including all four proposed social-monetisation features. This overlap provides additional evidence supporting the viability of social-monetisation features in enhancing the generalisability of fake news detection models. Notably, the ‘ads’ feature ranks highly in both datasets, reinforcing the idea that a key motivation for creating disinformation is often profit through advertising. This high ranking for `ads' aligns with findings in fake news literature that connect monetisation tactics, such as heavy ad placement, with disinformation. Additionally, the Facebook feature ranks highly in both datasets, indicating the prominent role social media platforms play in the dissemination of fake news. The consistent relevance of these features suggests that economic incentives, captured through social-monetisation indicators like advertisements and Facebook links, are significant drivers of disinformation. This aligns with prior research that highlights the exploitation of digital platforms for financial gain as a core characteristic of fake news.

From a stylistic perspective, exclamation marks consistently rank as the most important feature in both datasets, with ‘all caps’ words also ranking prominently. These features are frequently associated with fake news, particularly in sensationalist headlines or emotionally charged content. This emphasis on exclamations and capitalised words aligns with prior research that links these stylistic cues to disinformation. Additionally, features like ‘CD’ (cardinal numbers) and ‘single quotes’ also show high importance in both datasets, which could reflect the tendency of fake news content to use specific numbers or quotations for added emphasis or perceived authority.

These findings underscore the value of both social-monetisation and stylistic features in identifying fake news. The strong presence of social-monetisation features, combined with stylistic cues like exclamations and all-caps text, suggests that these elements are integral to creating and detecting disinformation. Together, they enhance model accuracy and contribute to the broader objective of building more generalisable fake news detection models.

\begin{table}[h]
\centering
\caption{Reduced Feature-Set Results}
\label{tab:study2-reducedresults}
\vspace{5pt}
\begin{adjustbox}{width=\textwidth}
\begin{tabular}{|l|c|c|c|c|c|c|c|c|c|c|}
\cline{2-11}
\multicolumn{1}{c|}{} & \multicolumn{5}{|c|}{\textbf{Original Feature-Set}} & \multicolumn{5}{c|}{\textbf{Reduced Feature-Set}} \\
\hline
\multicolumn{1}{|l|}{\textbf{Test}} & \textbf{Acc.} & \textbf{Prec.} & \textbf{Rec.} & \textbf{Spec.} & \textbf{F1} & \textbf{Acc.} & \textbf{Prec.} & \textbf{Rec.} & \textbf{Spec.} & \textbf{F1} \\
\hline
\textbf{K-Fold Test} & 0.94 & 0.93 & 0.96 & 0.92 & 0.94 & 0.91 & 0.89 & 0.94 & 0.89 & 0.91 \\
\textbf{Cross-Dataset Test} & 0.75 & 0.76 & 0.72 & 0.77 & 0.74 & 0.76 & 0.78 & 0.73 & 0.79 & 0.75 \\
\hline
\end{tabular}
\end{adjustbox}
\end{table}

Further analysis, involving the repetition of K-Fold cross-validation and cross-dataset testing using the 33 features that demonstrated positive feature importance across both datasets, revealed a slight decrease in K-Fold testing performance but slight improvements in cross-dataset testing on the Facebook URLs dataset. Specifically, accuracy, recall, and F1 score increased by 0.01, while precision and specificity each improved by 0.02. These findings indicate that the reduced feature set, while slightly compromising K-Fold testing performance, enhances generalisability when applied to external datasets. This underscores the value of prioritising features with consistent positive importance across datasets.

Compared to word embeddings such as Word2Vec and BERT, the reduced set of stylistic features offers notable advantages in terms of computational efficiency. Word embeddings typically require significant resources for both feature extraction and model training, particularly when fine-tuning pre-trained models on large datasets. In contrast, the streamlined stylistic feature set demands less computational overhead, enabling faster training and evaluation while maintaining competitive performance.

These results highlight the practical and efficient nature of stylistic features for real-world applications, where resource constraints and model scalability are critical considerations. By balancing performance, generalisability, and efficiency, the reduced feature set provides a compelling alternative to computationally intensive word embedding approaches.

\section{Discussion}
The motivation for this study stemmed from previous findings indicating poor generalisability of current fake news detection approaches. Current approaches often rely on token-representations and coarsely labelled datasets, that is, datasets that use the article’s publisher as a proxy for labelling articles as ‘fake’ or ‘true’. Building on findings that suggest stylistic features are less sensitive to biases in datasets, this study aimed to produce a generalisable model, trained on a coarsely labelled dataset (NELA) that performs well on real-world data (Facebook URLs Dataset). Additionally, the study proposed four social-monetisation features, and investigated the effectiveness of these novel features in producing more generalisable models

The first contribution of this study, relates to RQ1, confirming the issue of poor generalisability of models that use token-representations. Unlike previous studies that have examined generalisability across coarsely labelled datasets, this research focuses on 'real-world' data that has been manually fact-checked. Experiment 1 demonstrated that common token-representations (including BoW, TF-IDF, Word2Vec and BERT) suffer a similar drop in accuracy on the real-world data in the Facebook URLs dataset as other studies in the literature exploring generalisability between coarsely labelled datasets \citep{Silva2020,Lakshmanarao20193125,Kresnakova2019,Smitha2020}. There was also a large degree of variation in recall and specificity in these models, notably the BoW and TF-IDF Random Forest models and the BoW SVM model, which produced significantly higher values for recall compared to specificity. Additionally, the Word2Vec Logistic Regression model produced significantly higher specificity than recall. All these models provided similar levels of accuracy. Further exploration of LLMs in this study, specifically using the fine-tuned LLaMA model, revealed similar issues regarding generalisability and metric imbalances. The fine-tuned LLaMA model achieved an accuracy of 0.71 on the manually fact-checked Facebook URLs dataset, outperforming token-representation models in this regard. Additionally, it demonstrated high specificity (0.95), indicating strong performance in detecting fake news. However, recall was significantly lower (0.47) coupled with high precision (0.9), suggesting that while the model excelled at identifying fake news, it was conservative when classifying instances of true news. The zero-shot and few-shot configurations of LLaMA further highlighted its limitations in this domain. In the zero-shot configuration, the model performed poorly, failing to generalise effectively without task-specific fine-tuning. Similarly, the few-shot configuration, despite providing some task-specific context through labelled examples, showed substantial performance gaps, with accuracy and recall lagging behind those of the fine-tuned configuration. These results suggest that LLaMA, like other LLMs \citep{pavlyshenko2023analysis, kumar2024silver}, struggles to achieve robust performance in fake news detection with and without extensive adaptation to the task. This imbalance between specificity and recall across all LLaMA configurations mirrors the trends observed in token-representation models, underscoring the inherent difficulty in balancing these metrics across different approaches. These findings highlight the challenges of achieving generalisability and balanced performance when training models on coarsely labelled datasets commonly used throughout the literature.

Further reflection on the performance of these models also brings into focus an issue in current fake news research: there is, as yet, no consensus on whether to optimise for recall (accurate detection of true news) or specificity (accurate detection of fake news). It could be argued that optimising for recall is desirable as it may be critical to avoid censoring true news unintentionally, even if it is at the expense of misclassifying fake news. This approach prioritises legitimate information being freely disseminated, which is crucial in maintaining the integrity of open communication and the right to free speech. However, a high recall could lead to the spread of more false information, which can undermine public trust and have serious societal consequences Conversely, it could be argued that optimising for specificity is desirable, as capturing all instances of fake news is important, even if it is at the expense of misclassifying true news. This approach focuses on minimising the harm caused by misinformation, which can sway public opinion, impact elections, and incite unrest. However, excessive misclassification of true news could lead to censorship, limiting the diversity of viewpoints and potentially suppressing important information. Balancing these priorities is essential for developing effective and ethical fake news detection systems. As such, it is recommended that models aim to produce a similar ratio of false positives to false negatives, to ensure both classes of news are treated equally by the model.

The second contribution of this study pertains to RQ2, empirically validating that stylistic features can lead to more generalisable models. Experiment 2 provides evidence that while stylistic features did not significantly outperform token representations in terms of accuracy, they maintained more balanced recall and specificity. This balance suggests that models using stylistic features are less prone to producing false positives and false negatives, making them more suitable for generalisable fake news detection. Additionally, the resilience of stylistic features against potential biases related to specific topics and concept drift further contributes to their balanced performance \citep{Przybya2020}. As such, this may contribute to the model performing well across other datasets not utilised in this study. From a feature engineering and interpretability standpoint, stylistic features offer a straightforward and transparent means of identifying the specific features influencing the model (as demonstrated in the PFI analysis), unlike complex token representations \citep{Qiao2020}

The third contribution of this paper relates to the proposed social-\\monetisation features, used in addressing RQ3. These features, as demonstrated in Experiment 2, produced a statistically significant increase in accuracy under cross-dataset testing conditions on the manually labelled Facebook URLs dataset. The Random Forest and Gradient Boosting models, in particular, achieved a mean accuracy of 75\%, maintaining balanced specificity and recall. These results underscore the benefit of multimodal approaches that utilise features outside the article text, producing more robust and generalisable models. Similar to stylistic features, the proposed social-monetisation features can also be considered robust in regard to potential topical biases in datasets and concept drift. The feature importance analysis also demonstrates how these features, in particular the frequency of ads, positively contribute to producing more generalisable models.

The fourth contribution of this study relates to the PFI analysis, which demonstrated that a simplified feature-set performs comparably to the original comprehensive feature-set. Owing to the positive impact of the proposed social-monetisation features in the original model, this simplified model further underscores the utility of these novel features. The study also demonstrates the utility of this simplified feature-set alongside traditional machine learning algorithms such as Gradient Boosting. This simplified feature-set contributes to the efficiency in retraining of models, as well as extracting features for classifying unseen data, which is crucial for keeping pace with the constantly evolving news landscape. This is in contrast to fine-tuning large language models (LLMs), which can be time consuming and computationally expensive. 

Overall, this study makes significant contributions to the field of fake news detection by tackling the often-overlooked issue of generalisability. It highlights the limitations of training token-representations on coarsely labelled datasets, demonstrates the balanced performance provided by stylistic features, and introduces novel social-monetisation features that significantly improve model performance. These findings support the value of multimodal approaches in fake news detection and provides a foundation for future research to further enhance the robustness and applicability of fake news detection models in real-world scenarios

\section{Future Work and Conclusions}
While this study makes a significant contribution to the field of fake news detection by investigating the seldom addressed issue of generalisability using ‘real-world’ data, a number of limitations and therefore opportunities for future work have been identified.

Firstly, while this paper has explored the use of LLMs through the LLaMA model, providing initial insights into their application in this domain, future work may further investigate the potential of other fine-tuned LLMs to enhance generalisability and performance. Although this study, \cite{pavlyshenko2023analysis} and \cite{kumar2024silver} indicate that even advanced LLMs face challenges in achieving robust generalisability, continued research could examine integrating LLMs with other feature sets to address these limitations and improve performance in fake news detection tasks.

Secondly, although the inclusion of stylistic and social-monetisation features enhanced model performance and balance, the study was constrained to specific datasets. Future research should investigate the effectiveness of these features across a wider range of datasets, including those focused on different types of news topics, to better understand their generalisability and robustness across different domains. However, training and testing on coarsely labelled datasets can lead to misleading results that show high levels of performance in cross-validation or hold-out testing on unseen portions of training datasets, but not on real-world data. Given the limited availability of manually labelled real-world data like the Facebook URLs dataset used in this study, more effort is needed to produce granularly labelled datasets that can serve as robust benchmarks for evaluating fake news detection models. However, it is crucial to ensure user privacy and safety when creating these datasets, especially when derived from social media platforms such as Facebook or X/Twitter.

Moreover, while this study aimed to tackle the issue of poor generalisability from a feature engineering perspective, future work should focus on optimising model hyperparameters to further enhance the performance and robustness of fake news detection models, alongside the features proposed in this study. Fine-tuning hyperparameters such as learning rate, batch size and regularisation techniques could potentially improve model accuracy, recall, and specificity across different datasets and domains. Additionally, while promising results were produced in this study with the Gradient Boosting and Random Forest algorithms, it is also important to acknowledge the potential biases introduced by the models themselves. Algorithms such as Gradient Boosting prioritise features (such as exclamations and ads seen in the PFI analysis) that reduce the loss function. This prioritisation can potentially introduce biases if these features are not equally relevant across different datasets. Future work should therefore investigate other potential sources of bias, beyond dataset bias, to ensure the model's fairness and robustness. This includes a further examination of how features, especially the newly introduced 'social-monetisation' features, might inadvertently influence model predictions and contribute to biases. By addressing these biases, we can develop more reliable and equitable fake news detection models.

Finally, despite the comprehensive exploration of various stylistic features in this study, it is important to acknowledge that there are many other stylistic features that are yet to be explored in this context. Future work, therefore, should seek to identify other generalisable features, similar to the proposed social monetisation features, exclamations and all-caps words, as identified by this study. Additionally, given the advantages demonstrated through the four proposed novel features, future work should also try to identify such features that are available in the broader context of the whole webpage and not exclusively the article text. Further investigation into the computational efficiency of these features, compared to other approaches, should also be a priority in future research. This will ensure that the developed models can be efficiently deployed in real-time systems where computational resources and rapid response times are critical.

\bibliographystyle{apalike} 
\bibliography{cas-refs}

\newpage
\appendix
\section{Feature Tables}
\setcounter{table}{0}

\begin{table}[ht]
\centering
\caption{Fernandez Feature-Set}
\label{tab:fernandez}
\begin{adjustbox}{width=\textwidth}
\begin{tabular}{|l|p{12cm}|}
\hline
\textbf{Feature} & \textbf{Description} \\
\hline
Word Count & Total number of words \\
Syllables Count & Total number of syllables \\
Sentence Count & Total number of sentences \\
Word/Sent & Total words divided by total sentences \\
Long Words Count & Number of words with more than 6 characters \\
All Caps Count & Number of words in all caps \\
Unique Words Count & Number of unique words \\
Personal Pronouns \% & Percentage of words such as ‘I, we, she, him’ \\
First Person Singular \% & Percentage of words such as ‘I, me’ \\
First Person Plural \% & Percentage of words such as ‘we, us’ \\
Second Person \% & Percentage of words such as ‘you, your’ \\
Third Person Singular \% & Percentage of words such as ‘she, he, her, him’ \\
Third Person Plural \% & Percentage of words such as ‘they, them’ \\
Impersonal Pronouns \% & Percentage of words such as ‘it, that, anything’ \\
Articles \% & Percentage of words such as ‘a, an, the’ \\
Prepositions \% & Percentage of words such as ‘below, all, much’ \\
Auxiliary Verbs \% & Percentage of words such as ‘have, did, are’ \\
Common Adverbs \% & Percentage of words such as ‘just, usually, even’ \\
Conjunctions \% & Percentage of words such as ‘until, so, and, but’ \\
Negations \% & Percentage of words such as ‘no, never, not’ \\
Common Verbs \% & Percentage of words such as ‘run, walk, swim’ \\
Common Adjectives \% & Percentage of words such as ‘big, small, silly’ \\
Comparisons \% & Percentage of words such as ‘better, greater, larger’ \\
Concrete Figures \% & Percentage of words that represent real numbers \\
Punctuation Count & Total number of punctuation marks per document \\
Full Stop Count & Total number of full stops \\
Commas Count & Total number of commas \\
Colons Count & Total number of colons \\
Semi-Colons Count & Total number of semi-colons \\
Question Marks Count & Total number of question marks \\
Exclamation Marks Count & Total number of exclamation marks \\
Dashes Count & Total number of dashes \\
Apostrophe Count & Total number of apostrophes \\
Brackets Count & Total number of brackets ‘()’ \\
\hline
\end{tabular}
\end{adjustbox}
\end{table}

\begin{table}[ht]
\centering
\caption{Abonizio Feature-Set}
\label{tab:abonizio}
\begin{adjustbox}{width=\textwidth}
\begin{tabular}{|l|l|p{9cm}|}
\hline
\textbf{Group} & \textbf{Feature} & \textbf{Description} \\
\hline
\multirow{4}{*}{Complexity} & Word\_per\_sents & Average number of words per sentence \\
 & Avg\_word\_size & Average length of the words in the text \\
 & Sentences & Number of sentences \\
 & TTR & Type-Token Ratio – a metric of lexical variety \\
\hline
\multirow{17}{*}{Stylometric} & POS\_diversity\_ratio & Ratio of words with POS tags to length of text \\
 & Entities\_ratio & Ratio of named entities to length of text \\
 & Upper\_case & Number of upper-case letters \\
 & Oov\_ratio & Words that are OOV in Spacy’s language model \\
 & Quotes\_count & Number of quotation marks \\
 & Quotes\_ratio & Ratio of quotation marks to length of text \\
 & Ratio\_ADJ & Ratio of adjectives to text size \\
 & Ratio\_ADP & Ratio of adpositions to text size \\
 & Ratio\_ADV & Ratio of adverbs to text size \\
 & Ratio\_DET & Ratio of determiners to text size \\
 & Ratio\_NOUN & Ratio of nouns to text size \\
 & Ratio\_PRON & Ratio of pronouns to text size \\
 & Ratio\_PROPN & Ratio of proper nouns to text size \\
 & Ratio\_PUNCT & Ratio of punctuation to text size \\
 & Ratio\_SYM & Ratio of symbols to text size \\
 & Ratio\_VERB & Ratio of verbs to text size \\
\hline
\multirow{1}{*}{Psychological} & Polarity & Sentiment analysis score \\
\hline
\end{tabular}
\end{adjustbox}
\end{table}

\begin{table}[ht]
\centering
\caption{LIWC}
\label{tab:LIWC}
\begin{adjustbox}{width=\textwidth}
\begin{tabular}{|p{2.5cm}|p{5cm}|p{9cm}|}
\hline
\textbf{Group} & \textbf{Feature} & \textbf{Description} \\
\hline
Summary Variables 
& WC, Analytic, Clout, Authentic, Tone, WPS, BigWords, Dic 
& Word Count, Metric of logical/formal thinking, language of leadership/status, degree of +ve/-ve tone, average words per sentence, percentage of words \textgreater7 letters, percentage of words captured by LIWC dictionary \\
\hline
Punctuation Marks 
& Period, comma, qmark, exclam, apostro, otherp 
& Full stops, commas, question marks, exclamations, apostrophes, other punctuation \\
\hline
Linguistic Dimensions 
& Function, pronoun, ppron, I, we, you, shehe, they, ipron, det, article, number, prep, auxverb, adverb, conj, negate, verb, adj, quantity 
& Total function words, total pronouns, personal pronouns, personal pronouns (1st person singular), personal pronouns (1st person plural, singular), personal pronouns (2nd person), personal pronouns (3rd person singular), personal pronouns (3rd person plural), impersonal pronouns, determiners, articles, numbers, prepositions, auxiliary verbs, adverbs, conjunctions, negations, common verbs, common adjectives, quantities \\
\hline
Psychological Processes 
& Drives, affiliation, achieve, power, Cognition, allnone, cogproc, insight, cause, discrep, tentat, certitude, differ, memory, Affect, tone\_pos, tone\_neg, emotion, emo\_pos, emo\_neg, emo\_anx, emo\_anger, emo\_sad, swear, Social, secbehav, prosocial, polite, conflict, moral, comm, socrefs, family, friend, female, male 
& Drives, Affiliation, Achievement, Power, Cognition, All-or-none, Cognitive processes, Insight Causation, Discrepancy, Tentative, Certitude, Differentiation, Memory, Affect, Positive tone, Negative tone, Emotion, Positive emotion, Negative emotion, Anxiety, Anger Sadness, Swear words, Social processes, Social behaviour, Prosocial behaviour, Politeness, Interpersonal conflict, Moralization, Communication, Social referents, Family, Friends, Female references, Male references \\
\hline
Expanded Dictionary 
& Culture, politic, ethnicity, tech, lifestyle, leisure, home, work, money, relig, physical, health, illness, wellness, mental, substances, sexual, food, death, need, want, acquire, lack, fulfil, fatigue, reward, risk, curiosity, allure, perception, attention, motion, space, visual, auditory, feeling, time, focuspast, focuspresent, conversation, netspeak, assent, nonflu, filler 
& Words pertaining to the following categories: Culture, Politics, Ethnicity Technology, Lifestyle, Leisure, Home, Work, Money Religion, Physical, Health, Illness, Wellness, Mental health, Substances, Sexual, Food, Death, States, Need, Want, Acquire, Lack, Fulfilled, Fatigue, Motives, Reward, Risk, Curiosity, Allure, Perception, Attention, Motion, Space, Visual, Auditory, Feeling, Time orientation, Time, Past focus, Present focus, Future focus, Conversational, Netspeak, Assent, Nonfluencies, Fillers \\
\hline
\end{tabular}
\end{adjustbox}
\end{table}

\begin{table}[ht]
\centering
\caption{NELA Feature-Set}
\label{tab:NELA}
\begin{adjustbox}{width=1\textwidth}
\begin{tabular}{|p{5cm}|p{5cm}|p{9cm}|}
\hline
\textbf{Group} & \textbf{Feature} & \textbf{Description} \\
\hline
Style - Largely similar to those from the previous two studies, focusing on POS tags & 'quotes', ‘exclaim', ‘allpunc', 'allcaps', ‘stops', CC, CD, DT, EX, FW, IN, JJ, JJR, JJS, LS, MD, NN, NNS, NNP, NNPS, PDT, POS, PRP, PRP\$, RB, RBR, RBS, RP, SYM, TO, UH, VB, VBD, VBG, VBN, VBP, VBZ, WDT, WP, WP\$, WRB, ('\$',), ("''",), ('(',), (')',), (',',), ('--',), ('.',), (':',), ('``',) & Quotes, Exclamations, Punctuation Count, All Caps Count, Coordinating conjunction, Cardinal number, Determiner, Existential ‘there’, Foreign word, Preposition or subordinating conjunction, Adjective, Adjective (comparative), Adjective (superlative), List item marker, Modal, Noun (singular or mass), Noun (plural), Proper noun (singular), Proper noun (plural), Predeterminer, Possessive ending, Personal pronoun, Possessive pronoun, Adverb, Adverb (comparative), Adverb (superlative), Particle, Symbol, ‘to’, Interjection, Verb (base form), Verb (past tense), Verb (gerund or present participle), Verb (past participle), Verb (non-3rd person singular present), Verb (3rd person singular present), Wh-determiner, Wh-pronoun, Possessive wh-pronoun, Wh-adverb, Dollar signs, Double Quotations Marks, Open Parentheses, Closing Parentheses, Commas, Dashes, Sentence Terminators, Colons, Single Quotation Marks \\
\hline
Complexity - Assesses an article's complexity by analyzing lexical diversity, reading-difficulty metrics, and the average length of words and sentences. & 'ttr', 'avg\_wordlen', 'word\_count', 'flesch\_kincaid\_grade\_level', 'smog\_index', 'coleman\_liau\_index', 'lix' & Type-token ratio (variation of vocabulary), Average Word-Length, Word Count, Flesch Kincaid Grade (readability metric), SMOG Index (‘Simple Measure of Gobbledygook’), Coleman-Liau Index (readability metric), LIX (readability metric) \\
\hline
Bias - based on (Recasens et al., 2013), capture text subjectivity by identifying hedges, factives, assertives, implicatives, and opinion words. & 'bias\_words', 'assertatives', 'factives', 'hedges', 'implicatives', 'report\_verbs', 'positive\_opinion\_words', 'negative\_opinion\_words' & Bias words (word that introduce prejudice), assertatives (words stating facts with confidence), factives (words that imply truth), hedges (that determine the strength of a statement), implicatives (words that imply), report verbs (e.g., ‘report’ or ‘declare’), positive opinion words, negative opinion words \\
\hline
Affect - Relying on VADER sentiment analysis, this group aims to capture the emotion and sentiment of the text & 'vadneg', 'vadneu', 'vadpos', 'wneg', 'wpos', 'wneu', 'sneg', 'spos', 'sneu' & VADER Negative sentiment, VADER Neutral sentiment, VADER Positive Sentiment. The remaining tags refer to different types of words (positive, negative and neutral) that appear in a dictionary based on Recasens et al.’s work \\
\hline
Moral – Evaluates the ethical content of text using a lexicon developed from Moral Foundation Theory by Graham et al., further elaborated by Lin et al. & 'HarmVirtue', 'HarmVice', 'FairnessVirtue', 'FairnessVice', 'IngroupVirtue', 'IngroupVice', 'AuthorityVirtue', 'AuthorityVice', 'PurityVirtue', 'PurityVice', 'MoralityGeneral' & Words pertaining to the following categories: Caring for others, causing harm, fairness, unfairness, loyalty, disloyalty, authority, subversion, purity, degradation and general words in relation to morality \\
\hline
Event - Aims to capture words relating to dates, times and locations. & Num\_locations, num\_dates & Number of geographical locations, number of dates \\
\hline
\end{tabular}
\end{adjustbox}
\end{table}

\clearpage
\section{Feature Importance}
\setcounter{figure}{0}
\begin{figure}[p]
    \centering
    \includegraphics[width=0.7\linewidth,height=18cm]{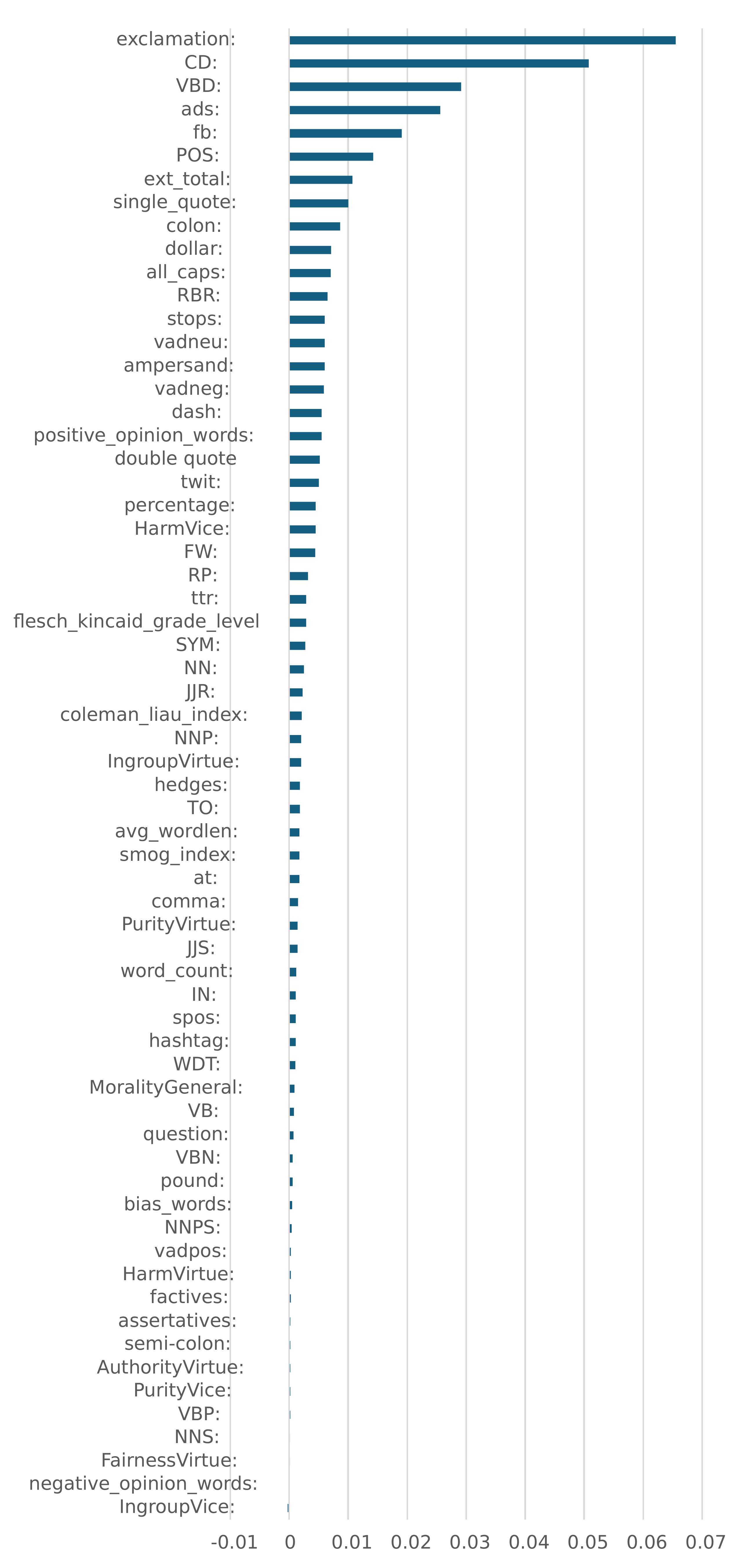}
    \caption{NELA Feature Importance}
    \label{fig:baseline-pfi}
\end{figure}

\begin{figure}[p]
    \centering
    \includegraphics[width=0.7\linewidth,height=18cm]{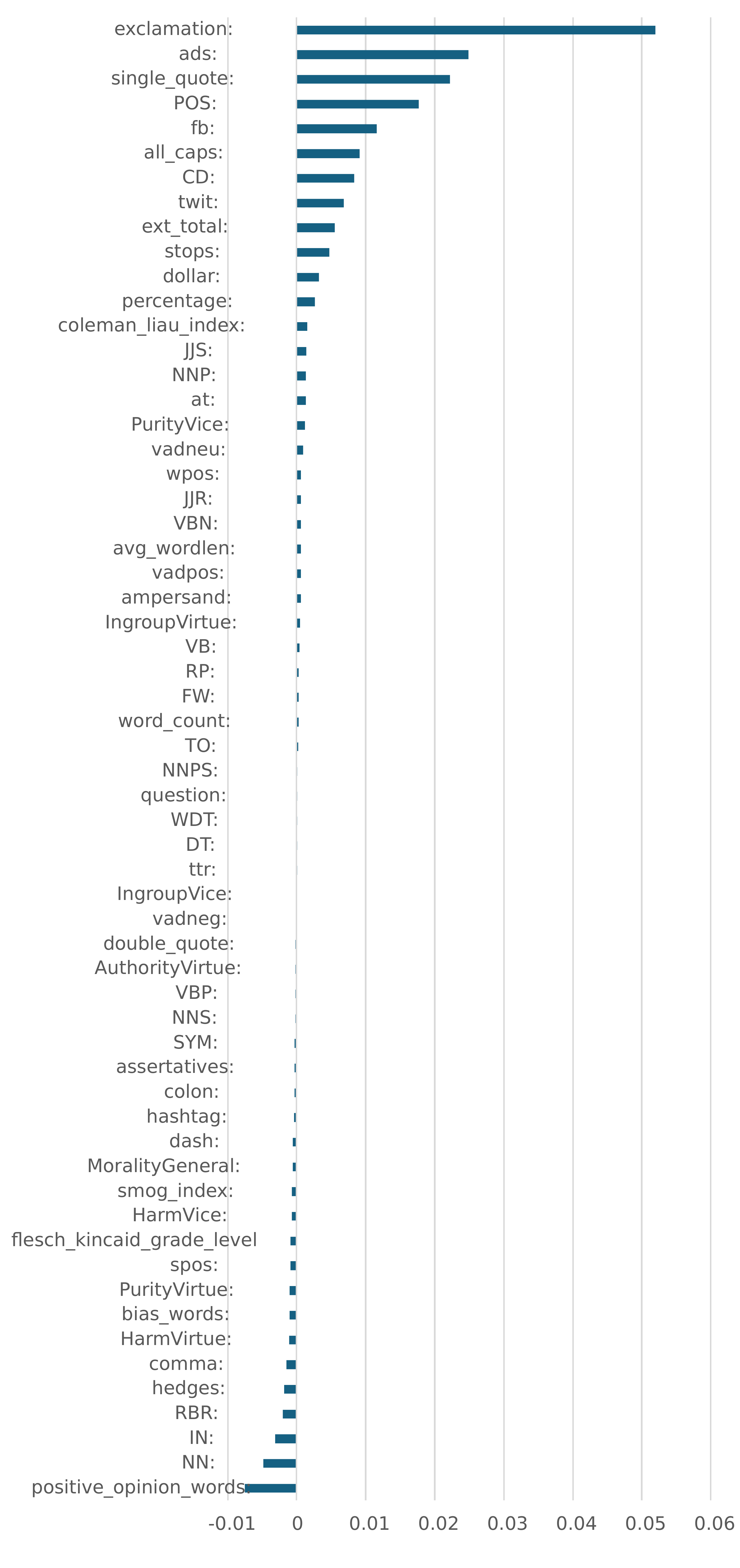}
    \caption{External Validation Feature Importance}
    \label{fig:external-pfi}
\end{figure}







\end{document}